%% file: acl_latex_arxiv.tex
\documentclass[11pt]{article}

\PassOptionsToPackage{table}{xcolor}
\usepackage[final]{acl}

\usepackage{times}
\usepackage{latexsym}
\usepackage{xurl}

\usepackage[T1]{fontenc}

\usepackage[utf8]{inputenc}

\usepackage{microtype}

\usepackage{inconsolata}

\usepackage{algorithm}
\usepackage{algorithmic}

\usepackage{amsfonts}
\usepackage{amssymb}
\usepackage{hyperref}
\usepackage{bm}
\usepackage{multirow}
\usepackage{multicol}

\usepackage{graphicx}
\usepackage{booktabs}
\usepackage{framed}
\usepackage{caption}
\usepackage{subcaption}
\usepackage{xspace}
\usepackage{enumitem}
\usepackage{xcolor}
\usepackage{color,colortbl}
\usepackage{amsthm}
\usepackage{amsmath}
\usepackage{tcolorbox}
\usepackage{cleveref}
\usepackage[inkscapelatex=false]{svg}
\usepackage{dashrule}
\usepackage{arydshln}
\usepackage{overpic}
\usepackage{marvosym}
\usepackage[table]{xcolor}
\usepackage{adjustbox}

\def\nsds{NSDS\xspace}
\def\seshort{SE\xspace}
\def\selong{Structural Expressiveness\xspace}
\def\selonglow{structural expressiveness\xspace}
\def\nvshort{NV\xspace}
\def\nvlong{Numerical Vulnerability\xspace}
\def\nvlonglow{numerical vulnerability\xspace}
\def\madsigmoid{MAD-Sigmoid\xspace}
\def\softor{Soft-OR\xspace}

\newcommand{\cdashlinelr}[1]{%
  \noalign{\vskip\aboverulesep
           \global\let\@dashdrawstore\adl@draw
           \global\let\adl@draw\adl@drawiv}
  \cdashline{#1}
  \noalign{\global\let\adl@draw\@dashdrawstore
           \vskip\belowrulesep}}

\makeatletter
\def\@fnsymbol#1{}
\makeatother

\newtheorem{insight}{Finding}

\tcbuselibrary{skins, breakable}

\definecolor{lightblue}{RGB}{220,235,250}

\newtcolorbox[auto counter]{SummaryBox}[1][]{ 
    enhanced,
    breakable,
    colback=lightblue!80,         
    colframe=black,               
    fonttitle=\bfseries\fontsize{10.1pt}{11.2pt}\selectfont,
    coltitle=white,               
    colbacktitle=black,           
    title={Key Finding \thetcbcounter}, 
    width=\linewidth,             
    arc=3.5mm,                    
    attach boxed title to top left={xshift=2.5mm, yshift=-2.5mm},
    boxed title style={rounded corners, size=small, colframe=black, colback=black},
    top=4mm, bottom=2mm, left=2mm, right=2mm,
}

\title{Beyond Outliers: A Data-Free Layer-wise Mixed-Precision Quantization Approach Driven by Numerical and Structural Dual-Sensitivity}

\author{
 \textbf{Hengyuan Zhang\textsuperscript{1}},
 \textbf{Xinrong Chen\textsuperscript{2}},
 \textbf{Zunhai Su\textsuperscript{1,3}},
 \textbf{Xiao Liang\textsuperscript{4}},
 \textbf{Jing Xiong\textsuperscript{1}},
 \textbf{Wendong Xu\textsuperscript{1}},\\
 \textbf{He Xiao\textsuperscript{1}},
 \textbf{Chaofan Tao\textsuperscript{1}}, 
 \textbf{Wei Zhang\textsuperscript{5}},
 \textbf{Ruobing Xie\textsuperscript{5}},
 \textbf{Lei Jiang\textsuperscript{5}},
 \textbf{Hayden Kwok-Hay So\textsuperscript{1}}, 
 \textbf{Ngai Wong\textsuperscript{1 \dag}}\thanks{\dag\ Corresponding author.}
\\
 \textsuperscript{1}The University of Hong Kong  \ \
 \textsuperscript{2}Peking University  \ \ 
 \textsuperscript{3}Tsinghua University  \ \ \\
 \textsuperscript{4}University of California, Los Angeles  \ \ 
 \textsuperscript{5}Tencent 
 \\
\texttt{hengyuan.zhang88@gmail.com }  
}

\begin{document}
\maketitle
\begin{abstract}
\input{abstract/abstract_arxiv}

\end{abstract}

\section{Introduction}
\label{sec:intro}
\input{intro/intro_arxiv}

\section{The NSDS Framework}
\label{sec:method}
\input{method/method_arxiv}

\section{Experiments}
\label{sec:exp}

\subsection{Experiment Settings}
\label{subsec:exp_setting}
\input{exp_setting/exp_setting}

\subsection{Performance of \nsds}
\label{sec:main_results}
\input{main_results/main_results_arxiv}

\vspace{-0.4cm}
\begin{figure}[!h]
    \centering \centerline{\includegraphics[width=\columnwidth]{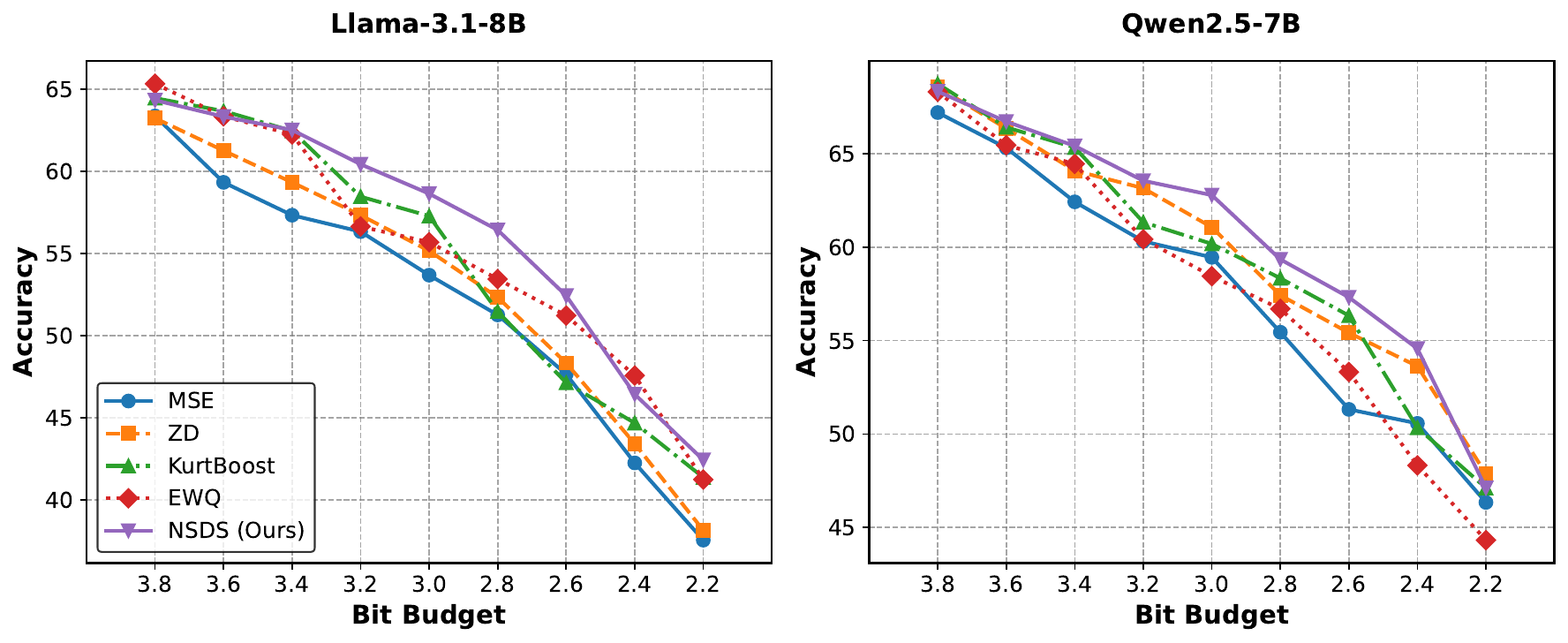}}
    \vspace{-0.2cm}
    \caption{Average accuracy of \nsds and baselines on language reasoning benchmarks across different bit budgets for Llama-3.1-8B and Qwen2.5-7B.}
    \vspace{-0.2cm}
    \label{fig:bit_budget_acc}
\end{figure}

\subsection{Further Analysis}
\label{sec:further_analysis}
\input{further_analysis/further_analysis_arxiv}

\begin{figure}[!th]
    \centering \centerline{\includegraphics[width=\columnwidth]{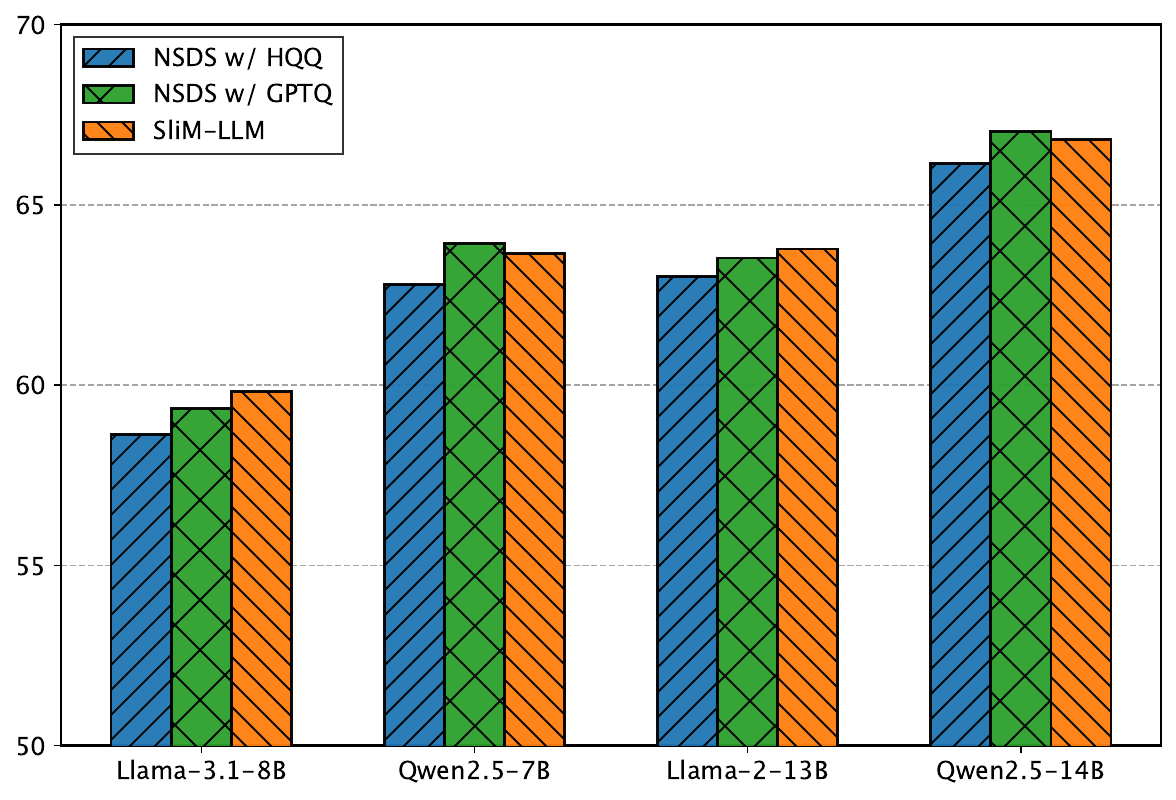}}
    \vspace{-0.2cm}
    \caption{Average accuracy on language reasoning benchmarks when integrating \nsds with different PTQ backends, compared against the SliM-LLM.}
    \vspace{-0.2cm}
\label{fig:integration_acc}
\end{figure}

\section{Related Work}
\label{sec:related_work}
\input{related_work/related_work}

\section{Conclusion}
\label{sec:conclusion}
\input{conclusion/conclusion}

\section*{Limitations}
\label{sec:limitation}
\input{limitation/limitation}

\bibliography{custom}

\newpage
\onecolumn
\twocolumn
\appendix
\input{appendix/appendix_main}

\end{document}

%% file: abstract/abstract_arxiv.tex
Layer-wise mixed-precision quantization (LM-PQ) enables effective compression under extreme low-bit settings by allocating higher precision to sensitive layers.
However, existing methods typically treat all intra-layer weight modules uniformly and rely on a single nu-merical property when estimating sensitivity, overlooking their distinct operational roles and structural characteristics.
To address this, we propose \nsds, a novel calibration-free LMPQ framework driven by \underline{N}umerical and \underline{S}tructural \underline{D}ual-\underline{S}ensitivity.
Specifically, it first mechanistically decomposes each layer into distinct operational roles and quantifies their sensitivity from both numerical and structural perspectives. 
These dual-aspect scores are then aggregated into a unified layer-wise metric through a robust aggregation scheme based on \madsigmoid and \softor to guide bit allocation. 
Extensive experiments demonstrate that \nsds consistently achieves superior performance compared to various baselines across diverse models and downstream tasks, without relying on any calibration data.
The code is available at \href{https://github.com/rattlesnakey/NSDS}{the provided link}.

%% file: intro/intro_arxiv.tex
Large Language Models (LLMs) have demonstrated remarkable performances across a wide range of applications, including complex reasoning~\citep{li2025system,liang2025sws,yu2025chain,liang2026divideconquer}, multilingual comprehension~\citep{multilingual_survey,zhang-etal-2025-shifcon,bu-etal-2025-alignx}, and various specialized domains~\citep{zhang2024balancing,financialtradingllm,zhang2025guilomo,zhang2025find,chang2025treereview}. 
However, their massive scale requires substantial computational and memory resources for deployment.
Post-Training Quantization (PTQ)~\citep{gptq} has emerged as a dominant compression technique, reducing memory footprint and inference cost while preserving model capabilities without the need for resource-intensive retraining.

\begin{figure}[!t]
    \centering \centerline{\includegraphics[width=\columnwidth]{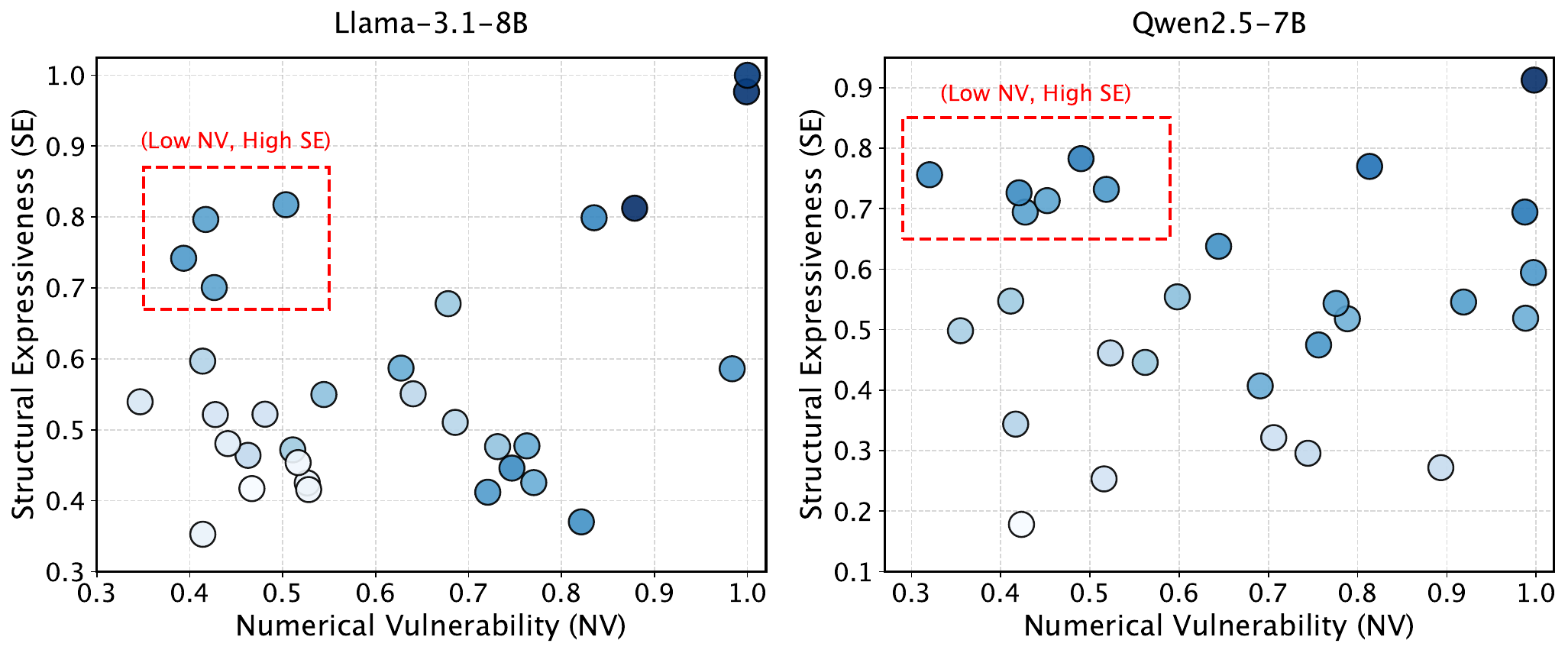}}
    \vspace{-0.2cm}
    \caption{Layer-wise sensitivity across two LLMs. Each dot represents a layer. Darker dots indicate more severe perplexity degradation ($\Delta$PPL) when quantizing that layer. See \S\ref{subsec:estimating_sensitivity} for the calculation of Numerical Vulnerability and Structural Expressiveness. Red dashed boxes highlight that layers with low numerical but high structural sensitivity still suffer significant degradation. See Appendix~\ref{app:nsds_scores} for more results.}
\vspace{-0.4cm}
    \label{fig:intro}
\end{figure}

Standard PTQ typically applies uniform precision across all layers, which often triggers severe performance degradation under extreme low-bit regimes (e.g., $<4$ bits)~\citep{llm-mq}. 
Layer-wise Mixed-Precision Quantization (LMPQ) alleviates this issue by strategically allocating higher precision to crucial layers and fewer bits elsewhere. 
Since bit-width remains uniform within each layer, LMPQ preserves tensor contiguity and kernel regularity, making it more hardware-friendly than other mixed-precision schemes~\citep{xiao_rc}.
Existing LMPQ methods are generally either search-based or criterion-based.
Since search-based optimization is computationally prohibitive and difficult to scale~\citep{haq,search1,amq}, criterion-based methods that efficiently estimate layer sensitivity via proxy metrics have become the preferred paradigm.
Depending on whether calibration data is required, they can be further divided into calibration-based and calibration-free methods. 
Calibration-based methods may overfit to the calibration set and require heavy forward and backward passes, which becomes increasingly impractical for large-scale LLMs~\citep{calibration_impact,kurtosis,intrinsicstructureproxysaliency}. 
Consequently, calibration-free methods that directly estimate sensitivity from model weights offer a more robust, efficient, and universally applicable solution for LMPQ.

Despite their efficiency, existing calibration-free LMPQ methods exhibit two major limitations: \textbf{1)} they typically treat all weight modules within a layer uniformly when estimating sensitivity, ignoring that different modules play distinct operational roles; \textbf{2)} they rely on a single numerical property, such as outlier statistics or parameter distribution, and do not consider the structural expressiveness of the weights. As shown in Fig.~\ref{fig:intro}, a layer with minimal numerical outliers can still suffer a significant performance drop after quantization if its structural expressiveness is compromised.

To address these limitations, we propose a calibration-free LMPQ approach driven by Numerical and Structural Dual-Sensitivity (\nsds) to comprehensively estimate layer sensitivity. 
Specifically, we first mechanistically decompose a model layer into distinct operational components, categorizing them into \textit{Detectors} and \textit{Writers}. 
We then quantify the sensitivity of each component from two perspectives: \nvlong (\nvshort) and \selong (\seshort). 
Finally, we aggregate these dual-aspect scores into a unified layer-wise metric using robust \madsigmoid normalization and \softor operations, and perform bit allocation based on the resulting ranking. 
Extensive experiments on various LLMs and benchmarks demonstrate the effectiveness of our \nsds framework.

To summarize, our contributions are as follows: 
\vspace{0.3em}

\quad \textbf{1)} We propose \nsds, a calibration-free LMPQ framework that estimates sensitivity from numerical and structural perspectives and integrates them through a robust aggregation scheme based on \madsigmoid and \softor, yielding a comprehensive criterion for bit allocation.

\vspace{0.3em}
\quad \textbf{2)} \nsds is the first to integrate the operational roles of weight components into quantization sensitivity estimation. We mechanistically decompose layers into operational components and design role-aware structural metrics for each component.

\vspace{0.3em}
\quad \textbf{3)} Extensive experiments validate the superiority of our \nsds framework. Without relying on any calibration data, \nsds consistently achieves competitive or superior performance compared to various baselines across different LLM families and downstream tasks. We also conduct further analysis to offer valuable insights for future research.

%% file: method/method_arxiv.tex
\begin{figure*}[!th]
    \centering \centerline{\includegraphics[width=2\columnwidth]{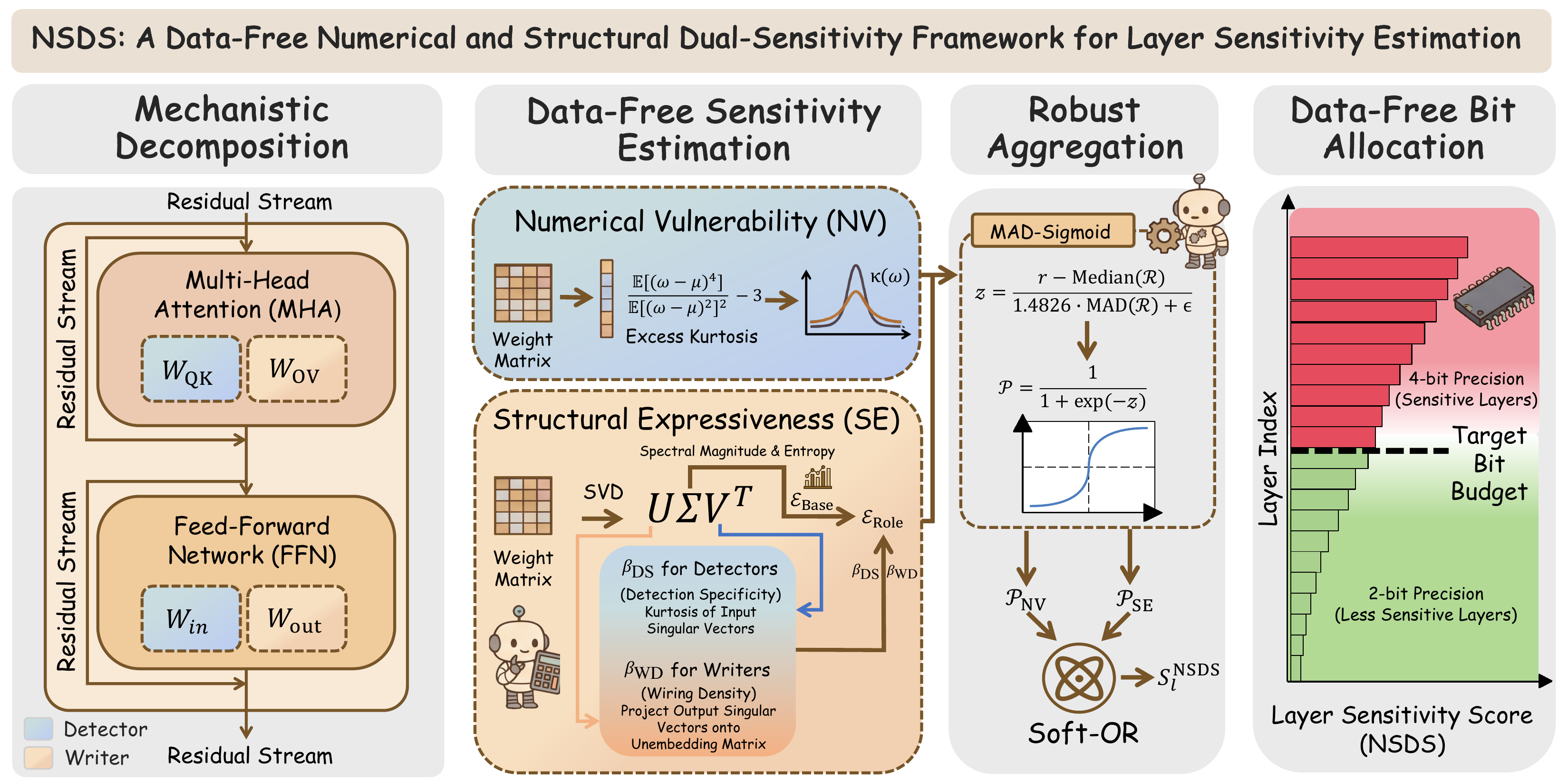}}
    \vspace{-0.1cm}
    \caption{Overview of our \nsds framework for data-free layer sensitivity estimation. The left panel illustrates the mechanistic decomposition of a layer into distinct operational components, categorized as \textit{Detectors} and \textit{Writers} (\S\ref{subsec:mechanistic_view}). The middle panel presents the dual-view sensitivity estimation: \nvlong (\nvshort) and \selong (\seshort) (\S\ref{subsec:estimating_sensitivity}). The right panel shows the robust aggregation process, where \madsigmoid normalizes heterogeneous scores and \softor integrates them into a unified layer-wise sensitivity metric, which is then used for bit allocation under a target budget (\S\ref{subsec:aggregation}). See Appendix~\ref{app:nsds_algo} for the algorithmic description of the \nsds framework.}
\vspace{-0.2cm}
    \label{fig:overview}
\end{figure*}

In this section, we introduce the \nsds framework, which comprehensively estimates layer sensitivity without requiring any calibration data. We first describe how a layer is mechanistically decomposed into distinct operational components (\S\ref{subsec:mechanistic_view}). Next, we present the metric to estimate the sensitivity from \nvlong and \selong (\S\ref{subsec:estimating_sensitivity}). 
Finally, we aggregate these scores into a unified layer-wise metric via \madsigmoid and \softor, and perform bit allocation based on the resulting ranking (\S\ref{subsec:aggregation}).
We present the overview of \nsds framework in Fig~\ref{fig:overview}.

\subsection{Mechanistic View of the Layer}
\label{subsec:mechanistic_view}

Previous methods typically treat all weight modules within a layer uniformly when estimating quantization sensitivity, overlooking the functional heterogeneity inherent in different components of the layer.
To address this, we decompose the layer based on mechanistic interpretability principles \citep{elhage2021mathematical,ferrando2024primerinnerworkingstransformerbased,zhang2026locatesteersurvey}. 
Typically, a layer consists of a Multi-Head Attention followed by a Feed-Forward Network.

\paragraph{Multi-Head Attention (MHA).} 
Given an input $X^{(l-1)} \in \mathbb{R}^{N \times d_{\text{model}}}$ from layer $l-1$, the MHA operation for head $h$ in layer $l$ is defined as:
\begin{equation}
\label{eq:mha_head}
\resizebox{\linewidth}{!}{%
$\displaystyle
\begin{aligned}
\mathrm{Attn}^{(l,h)}\bigl(X^{(l-1)}\bigr)
&= \operatorname{softmax}\!\Bigg(
    \frac{(X^{(l-1)}W_Q^{(l,h)})(X^{(l-1)}W_K^{(l,h)})^T}
         {\sqrt{d_{\mathrm{head}}}}
  \Bigg) \\
&\qquad\cdot\; X^{(l-1)}W_V^{(l,h)}W_O^{(l,h)}.
\end{aligned}$%
} 
\end{equation}
\noindent where $W_Q^{(l,h)}, W_K^{(l,h)}, W_V^{(l,h)}, W_O^{(l,h)}$ 
denote the query, key, value, and output projections for head $h$ in layer $l$.\footnote{For mechanistic decomposition, we equivalently split $W_O$ into per-head components $W_O^{(h)}$. See Appendix~\ref{app:split_output_proj} for details.}
Following \citet{elhage2021mathematical}, Eq.~\ref{eq:mha_head} can be rewritten to expose two distinct components:
\vspace{-0.2cm}
\begin{equation}
\resizebox{\linewidth}{!}{%
$\displaystyle
\begin{aligned}
\mathrm{Attn}^{(l,h)}(X^{(l-1)})
&= \operatorname{softmax}\!\left(
\frac{X^{(l-1)} \cdot 
\colorbox{red!15}{$W_Q^{(l,h)} W_K^{(l,h)T}$}
\cdot X^{(l-1)T}}
{\sqrt{d_{\mathrm{head}}}}
\right)
\\[4pt]
&\qquad \cdot\;
X^{(l-1)} \cdot
\colorbox{gray!15}{$W_V^{(l,h)} W_O^{(l,h)}$}
\\[8pt]
&= \operatorname{softmax}\!\left(
\frac{X^{(l-1)} W_{\mathrm{QK}}^{(l,h)} X^{(l-1)T}}
{\sqrt{d_{\mathrm{head}}}}
\right)
X^{(l-1)} W_{\mathrm{OV}}^{(l,h)}
\end{aligned}$%
}
\end{equation}
\noindent where $W_{\text{QK}}^{(l,h)} = W_Q^{(l,h)} W_K^{(l,h)T}$ and $W_{\text{OV}}^{(l,h)} = W_V^{(l,h)} W_O^{(l,h)}$, both $\in \mathbb{R}^{d_{\text{model}} \times d_{\text{model}}}$. 
The $W_{\text{QK}}^{(l,h)}$ component computes the attention pattern to \textit{detect} which tokens should be attended to.
The $W_{\text{OV}}^{(l,h)}$ component moves and transforms information from attended tokens, \textit{writing} it into the residual stream.

\paragraph{Feed-Forward Network (FFN).} The FFN takes the post-attention residual
$X_{\text{mid}}^{(l)} = X^{(l-1)} + \sum_{h=1}^H \mathrm{Attn}^{(l,h)}(X^{(l-1)})$
as input and is defined as:
\begin{equation}
    \text{FFN}^{(l)}(X_{\text{mid}}^{(l)}) = \sigma(X_{\text{mid}}^{(l)} W_{\text{in}}^{(l)}) W_{\text{out}}^{(l)}
\end{equation}
where $W_{\text{in}}^{(l)} \in \mathbb{R}^{d_{\text{model}} \times d_{\text{ffn}}}$, $W_{\text{out}}^{(l)} \in \mathbb{R}^{d_{\text{ffn}} \times d_{\text{model}}}$, and $\sigma(\cdot)$ is the activation function.\footnote{Modern SwiGLU-based LLMs include a gate projection $W_{\text{gate}}^{(l)} \in \mathbb{R}^{d_{\text{model}} \times d_{\text{ffn}}}$. See Appendix \ref{app:appendix_swiglu} for more details.} 
This operation can be viewed as \textit{key-value memory retrieval} \citep{geva2020transformer}.
As the keys, $W_{\text{in}}^{(l)}$ computes the neuron activation pattern, \textit{detecting} which neuron features are relevant to the input. 
As the values, $W_{\text{out}}^{(l)}$ \textit{writes} the activated neuron features back into residual stream according to their activation strengths.

\paragraph{Prediction as a Sum of Component Outputs.} 
To analyze the global effect of these components, we adopt the residual stream perspective where every component contributes via addition \citep{mickus2022prediction}. 
The unnormalized output logits $y$ for an $L$-layer model with $H$ attention heads can be expressed as:
\begin{equation}
\label{eq:info_flow}
\resizebox{\linewidth}{!}{%
$\displaystyle
y =
\left(
X^{(0)}
+
\sum_{l=1}^{L} \sum_{h=1}^{H}
\mathrm{Attn}^{(l,h)}(X^{(l-1)})
+
\sum_{l=1}^{L}
\mathrm{FFN}^{(l)}(X_{\mathrm{mid}}^{(l)})
\right)
W_U
$%
}
\end{equation}
where $X^{(0)}$ denotes the initial state of residual stream and $W_U \in \mathbb{R}^{d_{\text{model}} \times V}$ is the unembedding matrix for a vocabulary of size $V$. Based on this information flow, we categorize the components into two operational roles:
\begin{itemize}
    \item \textbf{Detector:} $W_{\text{QK}}^{(l,h)}$ and $W_{\text{in}}^{(l)}$ function as detectors.\footnote{For LLMs with $W_{\mathrm{gate}}^{(l)}$, we categorize $W_{\mathrm{gate}}^{(l)}$ as a detector. See Appendix~\ref{app:appendix_swiglu} for more details.} They compute attention or neuron activation patterns to detect related tokens or neuron features.
    \item \textbf{Writer:} $W_{\text{OV}}^{(l,h)}$ and $W_{\text{out}}^{(l)}$ function as writers. They process and write the detected information back into the residual stream, directly influencing the output logits in Eq.~\ref{eq:info_flow}.
\end{itemize}

\subsection{Estimating \nvlong (\nvshort) and \selong (\seshort)}
\label{subsec:estimating_sensitivity}
We comprehensively estimate layer sensitivity from two complementary perspectives: \nvshort and \seshort.

\paragraph{\nvlong (\nvshort).} 
We utilize \textit{excess kurtosis} to estimate the \nvlonglow of each component. 
Kurtosis measures the ``tailedness'' of a distribution. 
A larger kurtosis value indicates a heavy-tailed distribution with a higher density of extreme outliers. 
Quantizing components with such extreme outliers stretches the quantization range, easily triggering severe quantization degradation~\citep{xiao2023smoothquant, pham2025adaptive,su2026unveiling}. 
Given a weight matrix $W$ of a component, we first flatten it into a 1D vector $w$. The excess kurtosis $\kappa$ is then computed as:
\begin{equation}
    \kappa(w) = \frac{\mathbb{E}[(w - \mu)^4]}{(\mathbb{E}[(w - \mu)^2])^2} - 3
\end{equation}
where $\mu$ denotes the mean of flattened vector $w$.

\paragraph{\selong (\seshort).} 
We estimate \selonglow of each component by using its spectral magnitude and spectral entropy. 
Specifically, for a component $W$, we perform Singular Value Decomposition (SVD) such that $W = U \Sigma V^T$. 
Let $\boldsymbol{\sigma} = [\sigma_1, \sigma_2, \dots, \sigma_k]$ denote the vector of its singular values. 

The spectral magnitude, defined as the sum of the singular values ($||\boldsymbol{\sigma}||_1 = \sum_{i=1}^k \sigma_i$), captures the total structural energy. 
Meanwhile, the spectral entropy ($H$) captures the representational richness and information capacity~\citep{roy2007effective,xiong2025uncompmatrixentropyuncover}. 
To compute the spectral entropy, we first normalize the singular values into a distribution $p_i = \sigma_i / \sum_{j=1}^{k} \sigma_j$. The spectral entropy is then computed as the Shannon entropy:
\begin{equation}
    H(\boldsymbol{\sigma}) = - \sum_{i=1}^{k} p_i \log p_i
\end{equation}
Based on these theoretical properties, we formulate the base structural expressiveness ($\mathcal{E}_{\text{base}}$) as:
\begin{equation}
\label{eq:base_se}
    \mathcal{E}_{\text{base}} = ||\boldsymbol{\sigma}||_1 \times \exp(H(\boldsymbol{\sigma}))
\end{equation}

However, $\mathcal{E}_{\text{base}}$ is derived from raw singular values, reflecting intrinsic spectral properties of each component while overlooking its operational role.
A singular component with a small singular value may still play an important mechanistic role if it exhibits specific detection capabilities or targeted writing impacts. 
To bridge this gap, we reweight the raw singular values based on component's operational role.
We propose differentiated reweighting factors: \textit{Detection Specificity} ($\beta_{\text{DS}}$) for Detectors and \textit{Writing Density} ($\beta_{\text{WD}}$) for Writers.

\textbf{1) $\beta_{\text{DS}}$ for Detectors ($W_{\text{up}}$, $W_{\text{QK}}$).} 
Detectors rely on their input singular vectors ($V$) to detect patterns from the residual stream. 
To quantify this detection specificity, we use the excess kurtosis of these input singular vectors.
Specifically, for the $i$-th singular component, the reweighting factor $\beta_{\text{DS}}^{(i)}$ is computed as:
\begin{equation}
    \beta_{\text{DS}}^{(i)} = \kappa(v_i)
\end{equation}
where $v_i$ is the $i$-th column vector of $V$, and $\kappa(\cdot)$ denotes the excess kurtosis. 
A higher excess kurtosis in $v_i$ signifies that the detector is highly sharp and focuses on extremely specific patterns or features.

\textbf{2) $\beta_{\text{WD}}$ for Writers ($W_{\text{down}}$, $W_{\text{OV}}$).} 
Writers rely on their output singular vectors ($U$) to write the processed information into the residual stream.
Inspired by the logits lens framework~\citep{nostalgebraist2020,beren2022svd}, we quantify this writing density by projecting these output singular vectors onto the vocabulary unembedding matrix ($W_U$). 
Specifically, for the $i$-th singular component, the reweighting factor $\beta_{\text{WD}}^{(i)}$ is defined as the $L_1$ norm of this projection magnitude:
\begin{equation}
    \beta_{\text{WD}}^{(i)} = \| W_U^T u_i \|_1
\end{equation}
where $u_i$ is the $i$-th column vector of $U$.
A larger $\beta_{\text{WD}}^{(i)}$ indicates a stronger projection onto the vocabulary space, thereby exerting a greater impact on the final output distribution.

By element-wisely substituting the raw singular values $\sigma_i$ with their reweighted counterparts (i.e., $\sigma_i \leftarrow \sigma_i \cdot \beta_{\text{DS}}^{(i)}$ or $\sigma_i \leftarrow \sigma_i \cdot \beta_{\text{WD}}^{(i)}$) back into Eq.~\ref{eq:base_se}, we obtain the final role-aware structural expressiveness ($\mathcal{E}_{\text{role}}$) for each component.

\subsection{Score Aggregation and Bit Allocation}
\label{subsec:aggregation}
\paragraph{Robust Normalization via \madsigmoid.}

The estimated \nvshort and \seshort scores generally exhibit different scales and variances.
Direct aggregation would cause one metric to dominate due to scale mismatch. 
To align these metrics, we convert them into robust Z-scores~\citep{iglewicz1993volume} based on the Median Absolute Deviation (MAD).

Let $\mathcal{C} = \{W_{\mathrm{QK}}, W_{\mathrm{OV}}, W_{\mathrm{in}}, W_{\mathrm{out}}\}$ 
denote the set of component types in each layer, with $c \in \mathcal{C}$.
For a set of raw scores $\mathcal{R} = \{r^{(l,c)}\}_{l=1}^L$ (i.e., the SE or NV raw scores of component $c$ across all layers), we compute:
\begin{equation}
\label{eq:mad_sigmoid}
    z^{(l,c)} = \frac{r^{(l,c)} - \text{Median}(\mathcal{R})}{1.4826 \cdot \text{MAD}(\mathcal{R}) + \epsilon}
\end{equation}
where $\text{MAD}(\mathcal{R}) = \text{Median}(|r^{(l,c)} - \text{Median}(\mathcal{R})|)$, and $\epsilon > 0$ is a small constant for numerical stability. 
The factor $1.4826$ scales MAD to be comparable to a standard deviation under normality. 
We then map $z^{(l,c)}$ into the $(0, 1)$ range using the Sigmoid function to facilitate probabilistic-style aggregation: $\mathcal{P}^{(l,c)} = \frac{1}{1 + \exp(-z^{(l,c)})}$. 

\paragraph{Score Aggregation via \softor.}
We define the \softor operator as:
\begin{equation}
\text{\softor}\bigl(\{\mathcal{P}_i\}\bigr) 
= 
1 - \prod_i (1 - \mathcal{P}_i)
\end{equation}
For two terms, \softor reduces to 
\(
\mathcal{P}_1 + \mathcal{P}_2 - \mathcal{P}_1 \mathcal{P}_2,
\)
which increases whenever either term is large. 
Thus, \softor emphasizes the most sensitive components rather than averaging them.\footnote{To mitigate numerical saturation across $n$ components, here \softor is computed as $1 - \prod_{i=1}^{n} (1 - \mathcal{P}_i)^{\frac{1}{n}}$.}

Let $\kappa^{(l,c)}$ and $\mathcal{E}_{\text{role}}^{(l,c)}$ be the \nvshort and \seshort raw scores for component $c$ at layer $l$. 
The aggregation process proceeds as follows: 
\textbf{1)} Apply \madsigmoid to normalize $\kappa^{(l,c)}$ and $\mathcal{E}_{\text{role}}^{(l,c)}$ into $\mathcal{P}_{\text{\nvshort}}^{(l,c)}$ and $\mathcal{P}_{\text{\seshort}}^{(l,c)}$; 
\textbf{2)} Apply \softor across all components within layer $l$ to obtain the layer-wise sensitivities: $S_{l}^{\text{\nvshort}} = \text{\softor}\bigl(\{P_{\text{\nvshort}}^{(l,c)}\}_{c\in\mathcal{C}}\bigr)$ and $S_{l}^{\text{\seshort}} = \text{\softor}\bigl(\{P_{\text{\seshort}}^{(l,c)}\}_{c\in\mathcal{C}}\bigr)$;
\textbf{3)} Merge the layer-wise scores to derive the final \nsds metric:
\begin{equation}
    S_{l}^{\text{\nsds}} = \text{\softor}(S_{l}^{\text{\nvshort}}, S_{l}^{\text{\seshort}})
\end{equation}

\paragraph{Data-Free Layer-wise Bit Allocation.}
Given a target average-bit budget $\bar{b} \in [2, 4]$, the ratio of 4-bit layers under the assumption of equal-sized layers is $\rho = \frac{\bar{b}-2}{4-2}$. 
This yields a closed-form number of 4-bit layers $L_4 = \text{round}(\rho \cdot L)$, where $L$ is the total number of layers. Thus, we allocate 4-bit precision to $L_4$ most sensitive layers (i.e., with the highest $S_{l}^{\text{\nsds}}$), and assign 2-bit precision to the remaining $L_2 = L - L_4$ less sensitive layers.

%% file: exp_setting/exp_setting.tex
\paragraph{Evaluation Datasets and Metrics.}
We evaluate our method on both language modeling and language reasoning benchmarks. 
For language modeling, we report the Perplexity (PPL) on the WikiText-2~\citep{wikitext2} and C4~\citep{c4}.
For language reasoning, we measure the accuracy across ARC-Challenge~\citep{arc}, HellaSwag~\citep{hellaswag}, PIQA~\citep{piqa}, BoolQ~\citep{boolq}, WinoGrande~\citep{winogrande}, and TruthfulQA~\citep{truthfulqa}.

\paragraph{Models and Baselines.}
\label{para:model_baseline}
In our experiments, we employ models from the Llama-2 \citep{llama2}, Llama-3.1 \citep{llama3}, and Qwen2.5 \citep{qwen2.5} families, focusing on their 7B/8B and 13B/14B variants.
In the main experiments, we compare our \nsds with calibration-free layer-wise sensitivity metrics, including: MSE, EWQ~\citep{EWQ}, ZD~\citep{z_score_BI}, and KurtBoost~\citep{kurtosis}. 
For further analysis, we extend our comparison to several calibration-based layer-wise sensitivity metrics, including: LIM \citep{z_score_BI}, LSAQ \citep{LSAQ}, LLM-MQ \citep{llm-mq}, and LieQ \citep{xiao_rc}. 
Additionally, we compare our method against SliM-LLM \citep{slimllm}, a recent calibration-based group-wise mixed-precision method, to thoroughly demonstrate the efficacy of our framework.
See Appendix~\ref{app:baseline_details} for the details of baseline methods.

\paragraph{Implementation Details.}
In this work, we employ HQQ~\citep{hqq}, a calibration-free quantization method, as the default backend and conduct all evaluations under a target average-bit budget of $\bar{b}=3$, unless otherwise specified.
We set $\epsilon$ in Eq.~\ref{eq:mad_sigmoid} to $10^{-12}$.
For QK and OV components, NV and SE scores are first computed per head and then averaged across heads.
To filter out noise during the SE score computation (\S\ref{subsec:estimating_sensitivity}), we truncate all SVD matrices (including the unembedding matrix $W_U$) to retain only the top singular components that cumulatively account for 90\% of the total variance.
For settings that require calibration data, we randomly sample 128 sequences from Pile~\citep{pile}, each with 2048 tokens.
See Appendix~\ref{app:nsds_details} for more implementation details.

%% file: main_results/main_results_arxiv.tex
\input{tabs/main_results_tab}

Table~\ref{tab:main_results} reports the performance of the baseline methods and our proposed \nsds framework across various language reasoning and language modeling benchmarks for both Llama-3.1-8B and Qwen2.5-7B models. 
The results demonstrate that our \nsds consistently outperforms all baseline methods.

For instance, on Qwen2.5-7B, our \nsds surpasses the ZD baseline by 2.26\%, 1.58\%, and 3.07\% on ARC-C, Hellaswag, and PIQA, respectively.
Furthermore, relative to the baseline KurtBoost on Llama-3.1-8B, NSDS achieves accuracy gains of 3.08\% on ARC-C, 2.13\% on Hellaswag, and a more substantial 3.60\% on PIQA.
On language modeling tasks, \nsds also effectively minimizes the perplexity degradation, outperforming KurtBoost by 0.49 on Wikitext-2.
Based on these observations, we derive the following conclusion:
\begin{SummaryBox}
 \textit{By jointly considering numerical and structural sensitivity, our \nsds enables comprehensive sensitivity estimation, leading to more effective layer ranking.}
\end{SummaryBox}

%% file: tabs/main_results_tab.tex
\begin{table*}[!t]
    \centering
    
    \renewcommand{\arraystretch}{1.2} 
    
    \fontsize{10}{8}\selectfont
    
    \setlength{\tabcolsep}{5pt} 
    
    \begin{tabular}{l cccccc cc}
        \toprule[1.5pt]
        & \multicolumn{6}{c}{Language Reasoning ($\uparrow$)} & \multicolumn{2}{c}{Language Modeling ($\downarrow$)} \\
        \cmidrule(lr){2-7} \cmidrule(lr){8-9}
         & ARC-C & Hellaswag & PIQA & BoolQ & Winogrande & TruthfulQA & Wikitext-2 & C4 \\
        \midrule
        
        \rowcolor{gray!20} & \multicolumn{8}{c}{\textbf{Llama-3.1-8B}} \\ \addlinespace[1pt]
        FP16 & 57.76 & 81.97 & 80.09 & 82.11 & 77.35 & 28.40 & 6.24 & 8.95 \\ 
        \addlinespace[1pt] \hdashline[1pt/1pt] \addlinespace[2pt] 
        MSE & 38.73 & 64.21 & 68.58 & 60.73 & 65.82 & 23.95 & 9.22 & 12.35 \\
        EWQ & 40.17 & 67.16 & 71.49 & 64.52 & 67.43 & 23.34 & 8.38 & 11.27 \\
        ZD & 39.45 & 66.94 & 69.63 & 62.88 & 67.51 & 24.49 & 8.42 & 11.51 \\
        KurtBoost & 41.87 & 68.22 & 70.76 & \textbf{67.76} & 69.69 & 25.15 & 7.74 & 10.65 \\
        NSDS (Ours) & \textbf{43.16} & \textbf{69.67} & \textbf{73.31} & 66.89 & \textbf{72.28} & \textbf{26.43} & \textbf{7.25} & \textbf{9.97} \\
        \midrule

        \rowcolor{gray!20} & \multicolumn{8}{c}{\textbf{Qwen2.5-7B}} \\ \addlinespace[1pt]
        FP16 & 63.82 & 80.22 & 78.73 & 84.65 & 76.01 & 39.05 & 6.85 & 10.44 \\
        \addlinespace[1pt] \hdashline[1pt/1pt] \addlinespace[2pt] 
        MSE & 53.68 & 67.35 & 72.92 & 69.85 & 65.18 & 27.82 & 11.65 & 14.22 \\
        EWQ & 51.85 & 67.52 & 72.10 & 67.96 & 64.35 & 26.95 & 11.48 & 14.05 \\
        ZD & 55.82 & 70.45 & 73.01 & 72.95 & 65.25 & 28.94 & 9.23 & 12.15 \\
        KurtBoost & 54.25 & 69.92 & 72.48 & 70.40 & 65.72 & 28.35 & 9.84 & 13.68 \\
        NSDS (Ours) & \textbf{57.08} & \textbf{71.56} & \textbf{75.25} & \textbf{73.18} & \textbf{68.56} & \textbf{31.15} & \textbf{8.61} & \textbf{11.83} \\
        
        \bottomrule[1.5pt]
    \end{tabular}
    \vspace{-0.2cm}
    \caption{Results of baseline methods and our \nsds on Llama-3.1-8B and Qwen2.5-7B, evaluated on six language reasoning and two language modeling benchmarks. See Appendix~\ref{app:baseline_details} for the details of baseline methods.}
    \label{tab:main_results}
    \vspace{-0.2cm}
\end{table*}

%% file: further_analysis/further_analysis_arxiv.tex
\paragraph{\nsds Across Different Bit Budgets.}
\input{further_analysis/across_different_budget/across_different_budget_arxiv}

\paragraph{\nsds on Larger Model Scale.}
\input{further_analysis/larger_model_scale/larger_model_scale_arxiv}

\paragraph{Ablation Analysis of \nsds.}
\input{further_analysis/ablation_analysis/ablation_analysis_arxiv}

\paragraph{Compared with Calibration-based Methods.}
\input{further_analysis/compare_calibration_methods/compare_calibration_method_arxiv}

\paragraph{Integration with Other PTQ Backends.}
\input{further_analysis/integrate_other_quant/integrate_other_quant_arxiv}

%% file: further_analysis/across_different_budget/across_different_budget_arxiv.tex
We report the average accuracy of \nsds and baseline methods for Llama-3.1-8B and Qwen2.5-7B across different bit budgets in Fig.~\ref{fig:bit_budget_acc}.
As shown in Fig.~\ref{fig:bit_budget_acc}, all methods achieve similar performance at higher bit budgets. 
However, as the bit budget decreases, the baseline methods experience an early and sharp decline. 
For instance, the performance of EWQ and KurtBoost begins to degrade noticeably around 3.2 bits. In contrast, \nsds maintains a stable accuracy and does not show significant degradation until the bit budget drops to 2.6. 
Overall, we conclude that:
\begin{SummaryBox}
\textit{\nsds shows superior robustness across different bit budgets, maintaining high performance at lower bit budgets where baselines suffer severe degradation.}
\end{SummaryBox}

%% file: further_analysis/larger_model_scale/larger_model_scale_arxiv.tex
Having verified the effectiveness of our \nsds framework on 7B and 8B models, we further evaluate its scalability and generalization to larger-scale LLMs. 
As shown in Table~\ref{tab:larger_scale_results}, \nsds consistently outperforms all baseline methods in terms of both average PPL and average language reasoning accuracy. 
In particular, compared to the strong baseline KurtBoost, \nsds achieves an average accuracy improvement of 1.50\% and a PPL reduction of 0.40 on Llama-2-13B. 
Similarly, on the Qwen2.5-14B model, our method surpasses the KurtBoost baseline by 1.57\% in average accuracy and lowers the average PPL by 0.45. 
These findings demonstrate the robust scalability and broad effectiveness of our \nsds across different model scales and families.
\input{tabs/larger_scale_table}

%% file: tabs/larger_scale_table.tex
\begin{table}[!h] 
    \centering
    
    \renewcommand{\arraystretch}{1.2} 
    
    \fontsize{9}{7}\selectfont
    
    \setlength{\tabcolsep}{3pt} 
    
    \begin{tabular}{l cccc}
        \toprule[1.5pt]
        & \multicolumn{2}{c}{\textbf{Llama-2-13B}} & \multicolumn{2}{c}{\textbf{Qwen2.5-14B}} \\
        \cmidrule(lr){2-3} \cmidrule(lr){4-5}
        & Acc. ($\uparrow$) & PPL ($\downarrow$)  & Acc. ($\uparrow$) & PPL ($\downarrow$)  \\
        \midrule
        
        FP16        & 67.23 & 5.67 & 73.25 & 7.22 \\
        \addlinespace[1pt] \hdashline[1pt/1pt] \addlinespace[2pt] 
        MSE           & 59.91 & 9.38 & 64.29 & 10.62 \\ \addlinespace[1pt]
        EWQ    & 61.68 & 7.61 & 63.36 & 11.13 \\ 
        \addlinespace[1pt]
        ZD    & 60.99 & 7.98 & 65.42 & 10.25 \\
        \addlinespace[1pt]
        KurtBoost    & 62.08 & 7.51 & 65.13 & 10.41 \\ \addlinespace[1pt]
        \nsds (Ours) & \textbf{63.01} & \textbf{7.11} & \textbf{66.15} & \textbf{9.96} \\
        
        \bottomrule[1.5pt]
    \end{tabular}
    \vspace{-0.2cm}
    \caption{Average PPL and accuracy of \nsds and baselines on language modeling and reasoning benchmarks for larger-scale LLMs. See Table~\ref{tab:larger_main_results} for detailed results.} 
    \vspace{-0.4cm}
    \label{tab:larger_scale_results}
\end{table}

%% file: further_analysis/ablation_analysis/ablation_analysis_arxiv.tex
\begin{figure}[!th]
    \centering \centerline{\includegraphics[width=\columnwidth]{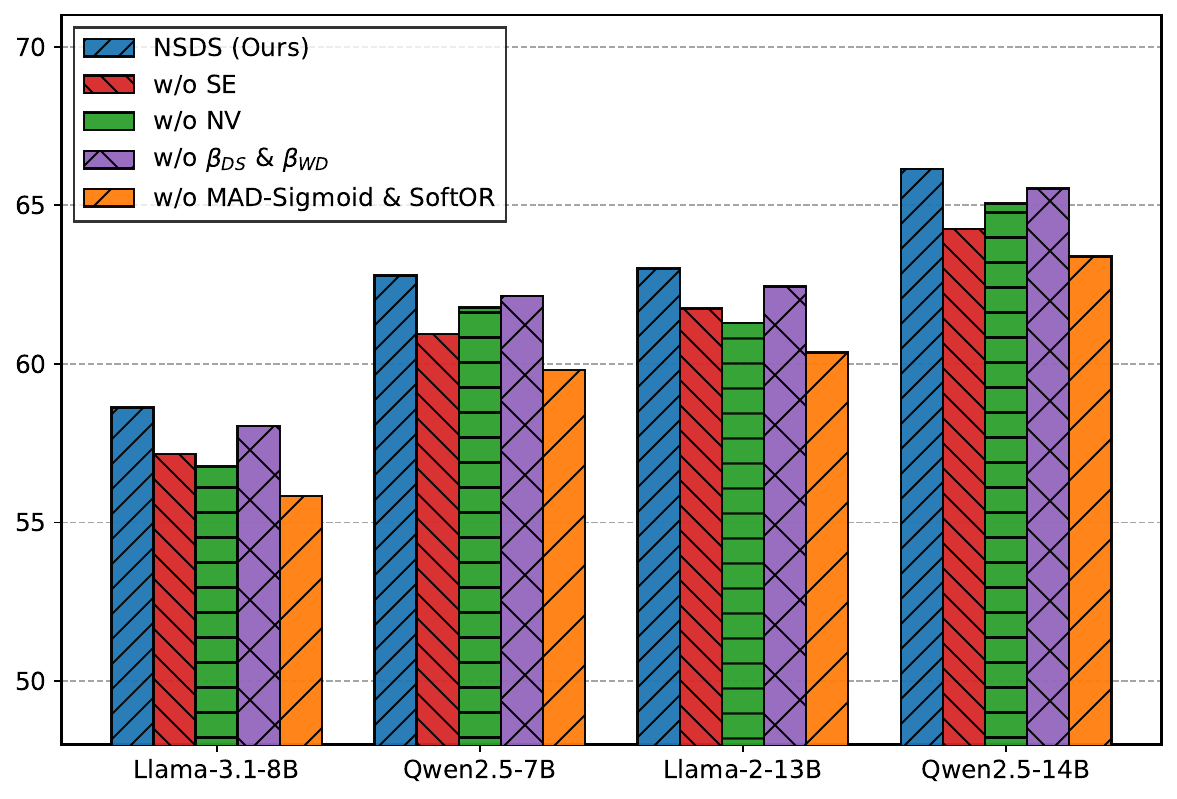}}
    \vspace{-0.3cm}
    \caption{Average accuracy of the ablation analysis on \nsds. ``w/o'' denotes the exclusion of a specific component from \nsds during layer sensitivity estimation.}
    \vspace{-0.4cm}
    \label{fig:ablation_accuracy}
\end{figure}

We conduct ablation analysis to examine the individual contributions of \nvshort, \seshort, the reweighting factors ($\beta_\text{DS}$ \& $\beta_\text{WD}$), and the aggregation operations (\madsigmoid \& \softor) within our \nsds framework. 
We report the average accuracy and PPL on language reasoning and language modeling benchmarks in Fig.~\ref{fig:ablation_accuracy} and Fig.~\ref{fig:ablation_ppl}, respectively.

As shown in Fig.~\ref{fig:ablation_accuracy}, the results demonstrate consistent performance drops across all models for all ablation variants. 
Notably, the performance drop is more pronounced when the aggregation functions are excluded (``w/o \madsigmoid \& \softor''), resulting in an accuracy drop of up to 4.08\% on Qwen2.5-14B. 
This highlights the crucial role of robust normalization and suitable aggregation operations when combining the SE and NV scores. 
Additionally, removing the reweighting factors (``w/o $\beta_\text{DS}$ \& $\beta_\text{WD}$'') causes a consistent accuracy drop, validating the effectiveness of role-aware singular value refinement.
These observations indicate that:
\begin{SummaryBox}
\textit{Proper normalization and aggregation are crucial for deriving layer-wise scores, and integrating NV with SE outperforms using either metric alone.}
\end{SummaryBox}

%% file: further_analysis/compare_calibration_methods/compare_calibration_method_arxiv.tex
\begin{figure}[!t]
    \centering \centerline{\includegraphics[width=\columnwidth]{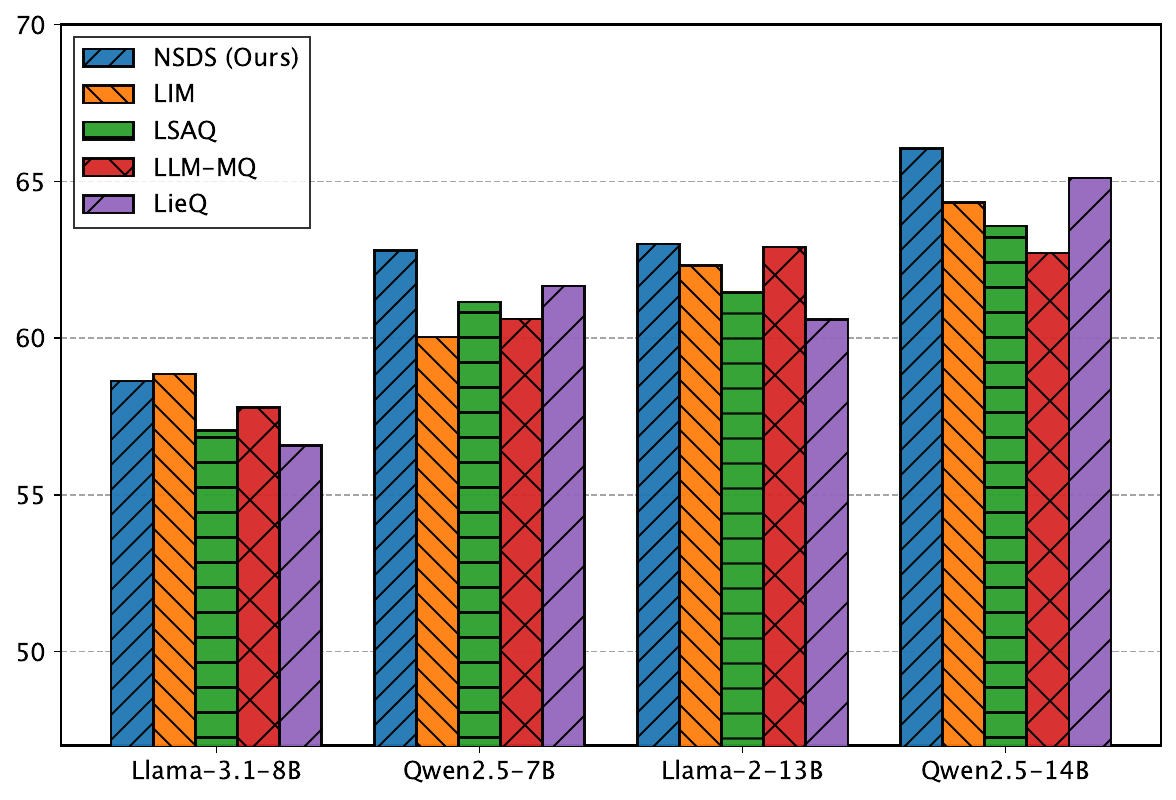}}
    \vspace{-0.3cm}
    \caption{Average accuracy comparison between \nsds framework and calibration-based baselines across generic reasoning benchmarks.}
    \vspace{-0.4cm}
\label{fig:calib_comparison_acc}
\end{figure}

We compare our data-free \nsds with several calibration-based layer sensitivity metrics, including LIM, LSAQ, LLM-MQ, and LieQ.
We report the average accuracy and PPL on language reasoning and language modeling benchmarks in Fig.~\ref{fig:calib_comparison_acc} and Fig.~\ref{fig:calib_comparison_ppl}, respectively.

As shown in Fig.~\ref{fig:calib_comparison_acc}, \nsds consistently ranks in the top two across all four evaluated LLMs, demonstrating cross-model robustness despite requiring no calibration data.
Specifically, \nsds achieves the highest average accuracy on three models (Qwen2.5-7B, Llama-2-13B, and Qwen2.5-14B), and ranks second on the Llama-3.1-8B, with only a marginal gap (within 0.49\%) behind the best-performing baseline.
In contrast, the calibration-based baselines exhibit notable performance fluctuations depending on the target model.
For instance, while LIM achieves the top rank on Llama-3.1-8B, its performance drops to the last place on Qwen2.5-7B.
Similarly, LLM-MQ performs the second best on Llama-2-13B but falls behind on Qwen2.5-14B.
This instability suggests that the effectiveness of calibration-based methods depends on how well the calibration data aligns with the specific target model, making them less robust across different models compared to \nsds.
Based on these observations, we derive the following conclusion:
\begin{SummaryBox}
\textit{Although fully data-free, our \nsds framework maintains strong cross-model robustness, whereas calibration-based methods lack such consistency.}
\end{SummaryBox}

%% file: further_analysis/integrate_other_quant/integrate_other_quant_arxiv.tex
Our \nsds is orthogonal to the choice of the PTQ backend. We replace the default HQQ backend with a stronger one, i.e., GPTQ, and compare it against SliM-LLM, a group-wise calibration-based mixed-precision method. 
We report the average accuracy and PPL in Fig.~\ref{fig:integration_acc} and Fig.~\ref{fig:integration_ppl}, respectively.
As shown in Fig.~\ref{fig:integration_acc}, when integrated with GPTQ, the performance of \nsds improves and shows comparable or even slightly superior performance to SliM-LLM.
Notably, SliM-LLM relies on calibration data and group-wise operations that are less hardware-friendly, whereas \nsds is calibration-free and adopts a hardware-friendly layer-wise scheme.
These observations indicate that:
\begin{SummaryBox}
\textit{Our \nsds is orthogonal to the underlying PTQ backend. It serves as a plug-and-play method where integrating a stronger quantization technique leads to better performance.}
\end{SummaryBox}

%% file: related_work/related_work.tex
\paragraph{Model Quantization.}
Quantization techniques for Large Language Models (LLMs) aim to reduce the precision of weights and activations to lower bits without significantly compromising model performance~\citep{zhu-etal-2024-survey-model,gong2025surveylowbitlargelanguage}. 
The two most prominent paradigms are Quantization-Aware Training (QAT)~\citep{liu-etal-2024-llm,chen-etal-2025-efficientqat} and Post-Training Quantization (PTQ)~\citep{10.5555/3454287.3455001,hqq,gptq,awq}. 
QAT incorporates quantization directly into the training phase, allowing the model to adapt to quantization errors. 
In contrast, PTQ applies quantization strictly after the model has been fully trained. 
As PTQ avoids the resource-intensive and time-consuming training processes required by QAT, it is more widely used in practice. 
In this paper, we focus on a specific class of PTQ known as weight-only quantization. 
This approach can be further categorized into calibration-based (e.g., GPTQ~\citep{gptq} and AWQ~\citep{awq}) and calibration-free methods (e.g., HQQ~\citep{hqq} and BnB~\citep{qlora}), depending on whether an extra calibration dataset is adopted.

\paragraph{Layer-wise Mixed-Precision Quantization (LM-PQ).}
Standard PTQ methods typically apply uniform bit allocation, often severely degrading performance under extreme low-bit regimes. 
LMPQ mitigates this by assigning higher precision to crucial layers and fewer bits elsewhere~\citep{kurtosis}.
As bit-width remains uniform within each layer, LMPQ preserves tensor contiguity and kernel regularity, making it more hardware-friendly than other mixed-precision schemes.
Prior LMPQ methods are either search-based or criterion-based. 
Search-based methods~\citep{haq,search1} are computationally intensive, making them impractical for many real applications. 
Conversely, criterion-based methods efficiently estimate layer sensitivity via proxy metrics to guide allocation.

Depending on whether calibration data is required, criterion-based methods can be further divided into calibration-based or calibration-free approaches. 
Calibration-based criteria leverage loss landscape information~\cite{hawqv2,llm-mq}, layer output discrepancies~\citep{z_score_BI,LSAQ}, or output representation compactness~\citep{xiao_rc}. 
However, they risk overfitting to calibration data and require resource-intensive forward and backward passes. 
Thus, we focus on the calibration-free paradigm. 
Existing calibration-free methods rely on heuristics like weight outliers~\citep{z_score_BI,kurtosis}, quantization errors~\cite{kloberdanz2023mixquantmixedprecisionquantization}, or parameter distribution entropy~\cite{EWQ}. 
However, they typically utilize single-faceted metrics focused solely on numerical magnitudes, ignoring the rich structural information of weights and treating all intra-layer modules uniformly.
To overcome these limitations, our NSDS framework mechanistically decomposes layers into distinct operational components, comprehensively estimating sensitivity from both numerical and structural perspectives.

%% file: conclusion/conclusion.tex
In this work, we propose \nsds, a novel calibration-free LMPQ framework that estimates layer sensitivity from both numerical and structural perspectives. 
Unlike previous calibration-free methods that uniformly treat all weight modules and rely solely on single numerical properties, \nsds mechanistically decompose each layer into distinct operational components and introduces role-aware sensitivity modeling that distinguishes between different functional modules within a layer. 
Through the integration of \madsigmoid normalization and \softor operation, \nsds produces a unified layer-wise ranking to guide bit allocation under a fixed budget.
Extensive experiments validate that \nsds consistently achieves competitive or superior performance compared to various baselines across different LLMs and downstream tasks, without relying on any calibration data. 
Our further analysis also offers valuable insights for future research.

%% file: limitation/limitation.tex
While our proposed \nsds framework demonstrates strong performance and robustness across multiple model families and benchmarks, our experiments are limited to models up to 14B parameters due to computational resource constraints. 
We have not evaluated \nsds on larger-scale models (e.g., 30B, 70B, or beyond), where sensitivity patterns may differ. 
Future work will evaluate \nsds on larger LLMs and additional model families to further investigate its scalability and generalization.

%% file: appendix/appendix_main.tex
\section{Visualization of \nsds Sensitivity Scores}
\label{app:nsds_scores}
\input{appendix/appendix_nsds_scores}

\section{Algorithmic Description of the \nsds Framework}
\label{app:nsds_algo}
\input{appendix/appendix_algo}

\section{Splitting the Output Projection Matrix}
\label{app:split_output_proj}
The original implementation of multi-head attention often considers a concatenation of each attention head output before projecting it into the global weight matrix $W_O \in \mathbb{R}^{H \cdot d_{\text{head}} \times d_{\text{model}}}$.
To properly decouple the components mechanistically, we split the concatenated $W_O$ into per-head weight matrices $W_O^{(h)} \in \mathbb{R}^{d_{\text{head}} \times d_{\text{model}}}$. 
By doing this, the individual matrices $W_V^{(h)}$ and $W_O^{(h)}$ can be mathematically joined into a single matrix $W_{\text{OV}}^{(h)}$ for each head $h$. 

\section{Implementation Details of the NSDS}
\label{app:nsds_details}
\input{appendix/appendix_methods}

\section{Details of Baselines}
\label{app:baseline_details}
\input{appendix/appendix_baseline}

\section{Details of PTQ Backends}
\label{app:details_ptq_backends}
\input{appendix/appendx_PTQ_backends}

\section{Additional Results in Further Analysis}
\label{app:further_analysis_results}
\input{appendix/appendix_further_analysis}

%% file: appendix/appendix_nsds_scores.tex
\begin{figure}[!th]
    \centering
    \includegraphics[width=\linewidth]{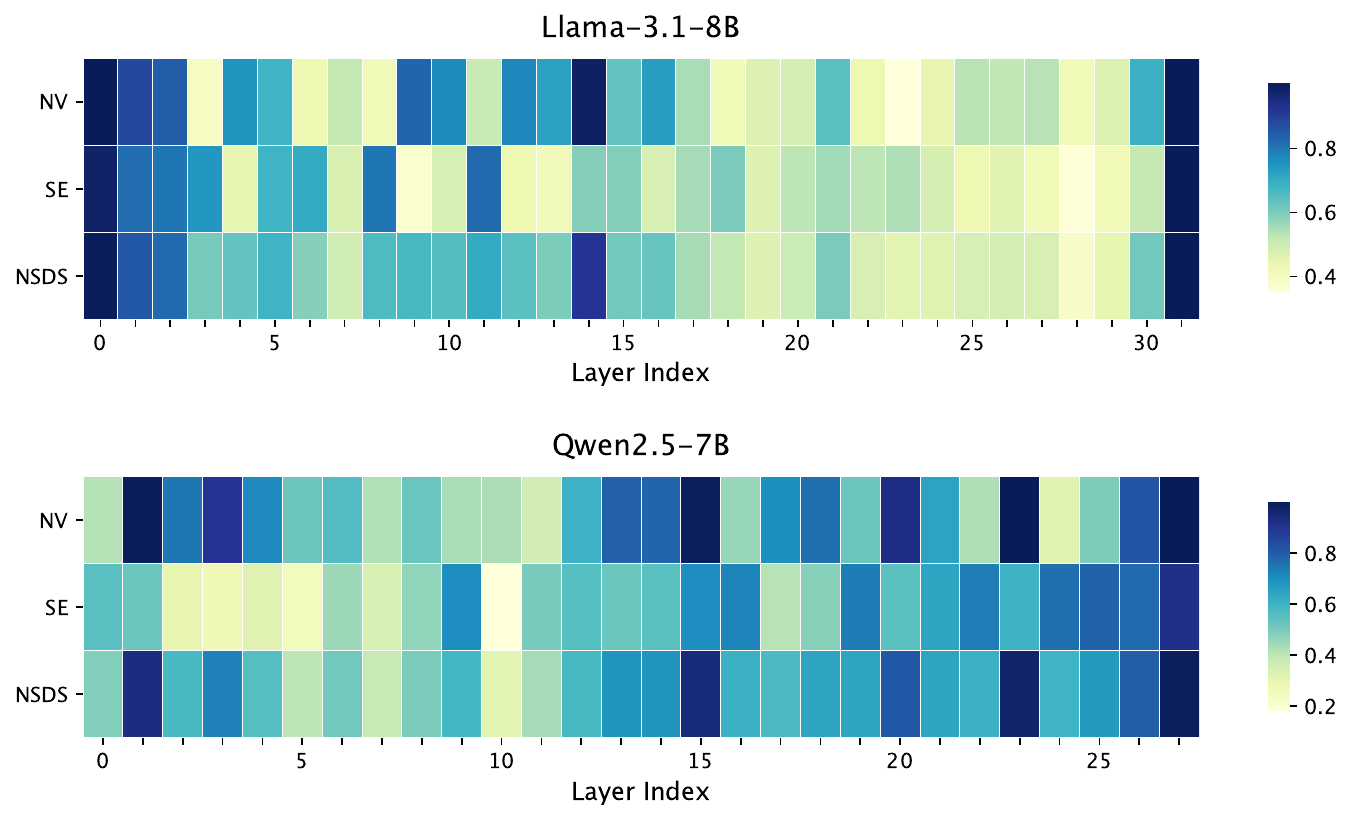} 
    \caption{Visualization of \nvshort, \seshort, and \nsds scores across layers on Llama-3.1-8B and Qwen2.5-7B. Darker colors indicate higher sensitivity score.}
    \label{fig:score_visualization}
\end{figure}

We present the visualization of \nvlong (\nvshort), \selong (\seshort), and our \nsds scores across layers on Llama-3.1-8B and Qwen2.5-7B in Fig.~\ref{fig:score_visualization}. 

As estimated by our \nsds metric, the layer-wise sensitivity exhibits distinct distribution patterns on Llama-3.1-8B and Qwen2.5-7B. Specifically, for Llama-3.1-8B, the most crucial and sensitive layers are predominantly located at the absolute bottom, the bottom-to-middle sections, and the topmost layers. 
Conversely, for Qwen2.5-7B, the high-sensitivity regions are primarily concentrated at the bottom and the middle-to-top sections. 

%% file: appendix/appendix_algo.tex
\begin{algorithm}[!tb]
\small
\caption{\nsds Framework for Data-Free Layer-wise Bit Allocation}
\label{alg:nsds_framework}
\begin{algorithmic}[1]
\REQUIRE Pre-trained LLM with $L$ layers, unembedding matrix $W_U$, component set $\mathcal{C} = \{W_{\mathrm{QK}}, W_{\mathrm{OV}}, W_{\mathrm{in}}, W_{\mathrm{out}}\}$, target average-bit budget $\bar{b} \in [2, 4]$.
\ENSURE Layer-wise bit allocation scheme $\mathcal{B} = \{b_1, b_2, \dots, b_L\}$.

\STATE \textbf{\% Phase 1: Metric Estimation (\nvshort and \seshort)}
\FOR{$l = 1$ \TO $L$}
    \FOR{each component $c \in \mathcal{C}$}
        \STATE Extract weight matrix $W^{(l,c)}$.
        \STATE Compute \nvshort score: $\kappa^{(l,c)} \leftarrow \text{Kurtosis}(W^{(l,c)})$.
        \STATE Perform SVD: $U, \Sigma, V^T \leftarrow \text{SVD}(W^{(l,c)})$ with singular values $\sigma_i$.
        \IF{$c \in \text{Detectors } (W_{\mathrm{QK}}, W_{\mathrm{in}})$}
            \STATE Compute reweighting factors: $\beta^{(i)} \leftarrow \kappa(v_i)$.
        \ELSIF{$c \in \text{Writers } (W_{\mathrm{OV}}, W_{\mathrm{out}})$}
            \STATE Compute reweighting factors: $\beta^{(i)} \leftarrow \| W_U^T u_i \|_1$.
        \ENDIF
        \STATE Update singular values: $\sigma_i \leftarrow \sigma_i \cdot \beta^{(i)}$.
        \STATE Compute role-aware \seshort score: $\mathcal{E}_{\text{role}}^{(l,c)} \leftarrow ||\boldsymbol{\sigma}||_1 \times \exp(H(\boldsymbol{\sigma}))$.
    \ENDFOR
\ENDFOR

\STATE \textbf{\% Phase 2: Robust Normalization and \softor Aggregation}
\STATE Normalize all $\kappa^{(l,c)}$ into $\mathcal{P}_{\text{\nvshort}}^{(l,c)}$ via \madsigmoid (Eq.~\ref{eq:mad_sigmoid}).
\STATE Normalize all $\mathcal{E}_{\text{role}}^{(l,c)}$ into $\mathcal{P}_{\text{\seshort}}^{(l,c)}$ via \madsigmoid (Eq.~\ref{eq:mad_sigmoid}).

\FOR{$l = 1$ \TO $L$}
    \STATE Aggregate \nvshort: $S_{l}^{\text{\nvshort}} \leftarrow 1 - \prod_{c \in \mathcal{C}} \left(1 - \mathcal{P}_{\text{\nvshort}}^{(l,c)}\right)^{\frac{1}{|\mathcal{C}|}}$
    \STATE Aggregate \seshort: $S_{l}^{\text{\seshort}} \leftarrow 1 - \prod_{c \in \mathcal{C}} \left(1 - \mathcal{P}_{\text{\seshort}}^{(l,c)}\right)^{\frac{1}{|\mathcal{C}|}}$
    \STATE Compute \nsds score: $S_{l}^{\text{\nsds}} \leftarrow S_{l}^{\text{\nvshort}} + S_{l}^{\text{\seshort}} - S_{l}^{\text{\nvshort}} \cdot S_{l}^{\text{\seshort}}$
\ENDFOR

\STATE \textbf{\% Phase 3: Data-Free Bit Allocation}
\STATE Calculate number of 4-bit layers: $L_4 \leftarrow \text{round} \left( \frac{\bar{b}-2}{2} \cdot L \right)$.
\STATE Sort all layers in descending order based on $S_{l}^{\text{\nsds}}$.
\STATE Assign $b_l = 4$ to the top $L_4$ layers, and $b_l = 2$ to the remaining layers.
\RETURN $\mathcal{B} = \{b_1, b_2, \dots, b_L\}$.
\end{algorithmic}
\end{algorithm}

To provide a clear and structured overview of our method, we summarize the entire \nsds framework in Algorithm~\ref{alg:nsds_framework}. 
The procedure consists of three main phases. 
First, we iterate through all layers and mechanistically decompose each layer into distinct components (Detectors and Writers). 
For each component, we compute its \nvlong (\nvshort) using excess kurtosis, and its \selong (\seshort) through role-aware spectral capacity. 
Second, to ensure scale-invariant aggregation, we apply a robust \madsigmoid normalization, followed by a \softor operation. 
This operation aggregates the component-level metrics into a unified layer-wise sensitivity score, emphasizing the most vulnerable parts of the layer without suffering from numerical saturation. 
Finally, based on the aggregated \nsds scores, we perform a deterministic, data-free bit allocation to strictly satisfy the target average-bit budget, prioritizing higher precision for the most sensitive layers.

%% file: appendix/appendix_methods.tex
In \S\ref{sec:method}, we introduced the high-level formulation of the proposed \nsds metric. 
This appendix provides further technical details of the \nsds framework.

\subsection{SwiGLU and Gate Projection}
\label{app:appendix_swiglu}
Modern Large Language Models (e.g., Llama, Qwen) frequently replace standard activation functions with SwiGLU. In this architecture, the Feed-Forward Network contains an additional gate projection weight $W_{\text{gate}}$. The operation is defined as:
\begin{equation}
    \text{FFN}(X) = \left( \text{Swish}(X W_{\text{gate}}) \odot X W_{\text{in}} \right) W_{\text{out}}
\end{equation}
Mechanistically, $W_{\text{gate}}$ reads from the residual stream and computes a gating distribution. It acts as an informational valve, \textit{detecting} whether specific features should be activated. Because its role is to compute a distribution that scales the hidden features, we categorize $W_{\text{gate}}$ as a \textbf{Detector}, identical to the role of $W_{\text{in}}$ and $W_\text{QK}$.

\subsection{Handling Grouped Query Attention}
To reduce memory overhead, many recent LLMs employ Grouped Query Attention (GQA)~\citep{GQA}, where multiple query heads share a single key and value head. When computing the $W_{\text{QK}}$ and $W_{\text{OV}}$ components for models utilizing GQA, we replicate (broadcast) the shared Key ($W_K$) and Value ($W_V$) matrices across their corresponding Query ($W_Q$) and Output ($W_O$) heads. This ensures that every attention head $h$ possesses independent, fully formed $W_{\text{QK}}^{(h)}$ and $W_{\text{OV}}^{(h)}$ matrices for computing \nvshort and \seshort scores in \S\ref{subsec:estimating_sensitivity}.

\subsection{SVD Top-90\% Variance Truncation}
When computing SE score in \S\ref{subsec:estimating_sensitivity}, incorporating the entire spectrum of singular values introduces long-tail noise. To capture the most principal structural features, we truncate the SVD matrices $U$ and $V$, as well as the singular values $\Sigma$, to retain only the top components that cumulatively account for 90\% of the total variance (energy). All subsequent calculations for $\mathcal{E}_\text{role}$ also operate on this truncated Top-90\% subspace. Similarly, for unembedding matrix $W_U$ used in $\beta_{\text{WD}}^{(i)}$, we pre-compute its Top-90\% SVD subspace to filter out vocabulary noise.

\subsection{Robust Sub-Linear Reweighting}
When reweighting the singular values $\Sigma$ using factor $\beta_{\text{DS}}^{(i)}$, extreme outliers in $\beta_{\text{DS}}^{(i)}$ can cause an exponential explosion, completely overshadowing the base structural capacity. To prevent mathematical collapse and filter out negative noise, we apply a robust sub-linear transformation to the raw reweighting factor $x$:
\begin{equation}
    \beta_{\text{DS}}^{(i)} = \log(1 + \text{ReLU}(x))
\end{equation}
The $\text{ReLU}$ operation elegantly sets the weight of completely flat or uniform distributions (where kurtosis $< 0$) to zero. The $\log(1 + \cdot)$ operation provides sub-linear scaling, ensuring that highly specific detectors are rewarded adequately without triggering numerical instability.

\subsection{Specifics of the QK Component Calculation}
While the up-projection ($W_\text{up}$) acts as a unidirectional feature reader relying solely on the input singular vectors ($V$), the QK circuit computes the interaction between Queries and Keys. Therefore, a highly specific attention routing requires both the Query vectors ($U$) and the Key vectors ($V$) to be sharp. In our implementation of the $\beta_{\text{DS}}^{(i)}$ for the QK component, the reweighting factor is computed as the product of their respective excess kurtosis values: $\beta_{\text{DS}}^{(i)} = \kappa(v_i) \times \kappa(u_i)$.

%% file: appendix/appendix_baseline.tex
In this Appendix, we introduce the details of the compared baselines involved in our experiments.

\subsection{Calibration-free Layer-wise Sensitivity Metric}
\paragraph{MSE.} 
The Mean Squared Error (MSE) metric estimates layer sensitivity by measuring the total numerical deviation between the full-precision weights and their dequantized (reconstructed) counterparts. To obtain the layer-level sensitivity score, we compute the sum of squared errors across all weight matrices within the layer. 
Let $\mathcal{W}_l$ denote the set of all weight matrices in layer $l$. The sensitivity score is calculated as:
\begin{equation}
    \mathcal{S}_{\text{MSE}}^{(l)} = \sum_{W \in \mathcal{W}_l} \| W - \hat{W} \|_F^2
\end{equation}
where $W$ and $\hat{W}$ are the full-precision and dequantized weight matrices, respectively. A larger $\mathcal{S}_{\text{MSE}}^{(l)}$ indicates a higher total reconstruction error, signifying that the layer is more sensitive and crucial to preserve.

\paragraph{ZD.} 
The Z-score Distribution (ZD) metric estimates layer sensitivity by analyzing the statistical distribution of parameter weights, notably requiring no calibration data. Specifically, it quantifies the importance of a layer based on the proportion of its weights that are significantly larger than the layer's average. 
Let $\mathcal{W}_l$ denote the set of all weight matrices in layer $l$, and let $N_l$ be the total number of scalar weight parameters in this layer. We first compute the mean $\mu_l$ and standard deviation $\sigma_l$ of all weights $w \in \mathcal{W}_l$. The Z-score for each individual weight $w$ is defined as:
\begin{equation}
    z = \frac{w - \mu_l}{\sigma_l}
\end{equation}
The ZD sensitivity score for layer $l$ is then calculated as the fraction of weights whose Z-score strictly exceeds $1$:
\begin{equation}
    \mathcal{S}_{\text{ZD}}^{(l)} = \frac{1}{N_l} \sum_{W \in \mathcal{W}_l} \sum_{w \in W} \mathbb{I}(z > 1)
\end{equation}
where $\mathbb{I}(\cdot)$ is the indicator function. According to the original formulation, a smaller $\mathcal{S}_{\text{ZD}}^{(l)}$ score implies a higher layer sensitivity, thus prioritizing the layer for higher-precision quantization.

\paragraph{EWQ.}
Entropy-Weighted Quantization (EWQ) estimates layer sensitivity by analyzing the entropy of the weight distributions. 
For a weight matrix $W \in \mathcal{W}_l$ with $N$ elements $\{w_i\}_{i=1}^N$, EWQ first flattens and normalizes the weights into a probability distribution using the softmax function: $p_i = \frac{\exp(w_i)}{\sum_{j=1}^N \exp(w_j)}$. 
The entropy of the weight matrix, denoted as $H(W)$, is then computed as:
\begin{equation}
H(W) = -\sum_{i=1}^N p_i \log(p_i + \epsilon)
\end{equation}
where $\epsilon$ is a small constant (e.g., $0.01$) added for numerical stability.

To obtain the layer-level sensitivity score, EWQ calculates the parameter-weighted average entropy across all weight matrices within the layer. 
Let $|W| = N$ denote the total number of parameters in the weight matrix $W$. The sensitivity score for layer $l$ is defined as:
\begin{equation}
\mathcal{S}_{\text{EWQ}}^{(l)} = \frac{\sum_{W \in \mathcal{W}_l} |W| H(W)}{\sum_{W \in \mathcal{W}_l} |W|}
\end{equation}

A larger $\mathcal{S}_{\text{EWQ}}^{(l)}$ indicates that the layer contains higher variability and captures more complex global information, making it more sensitive to quantization.

\paragraph{KurtBoost.} 
KurtBoost estimates layer sensitivity by analyzing the weight distribution's kurtosis. For a weight matrix $W \in \mathcal{W}_l$ with $N$ elements $\{w_i\}_{i=1}^N$, the raw kurtosis is computed as:
\begin{equation}
    k(W) = \frac{\frac{1}{N}\sum_{i=1}^{N}(w_i - \overline{W})^4}{\left(\frac{1}{N}\sum_{i=1}^{N}(w_i - \overline{W})^2\right)^2}
\end{equation}
where $\overline{W}$ is the mean of the weights in $W$. Unlike our NSDS framework which utilizes excess kurtosis (subtracting $3$), KurtBoost uses the raw kurtosis directly. 
The layer-wise sensitivity score $k^{(l)}$ is computed by: $k^{(l)} = \frac{1}{|\mathcal{W}_l|} \sum_{W \in \mathcal{W}_l} k(W)$.

To identify the most sensitive layers, KurtBoost constructs a difference sequence between adjacent layers, defined as $D = \{d_l\} = \{ k^{(l+1)} - k^{(l)} \}$ (or alternatively $k^{(l+1)}/k^{(l)}$ for monotonically increasing patterns). Assuming the difference set $D$ follows a normal distribution, a Z-score is computed for each variation:
\begin{equation}
    z_l = \frac{|d_l - \mu|}{\sigma}
\end{equation}
where $\mu$ and $\sigma$ are the mean and standard deviation of $D$. Layers with significant variations (i.e., $z_l > 3$) are flagged as outliers. During bit allocation, these detected outlier layers are strictly prioritized for high-precision assignment. Once the outliers are preserved, the remaining layers are selected and allocated higher precision based on their descending $k^{(l)}$ score ranking until the target bit budget is exhausted.

\subsection{Calibration-based Layer-wise Sensitivity Metrics}

\paragraph{LIM.} 
The Layer Importance Metric (LIM) estimates layer sensitivity by calculating the cosine similarity between the layer's input and output representations. Using a set of calibration data, let $X_{\text{in}}^{(l)}$ and $X_{\text{out}}^{(l)}$ denote the input and output hidden states of layer $l$, respectively. The sensitivity score is formulated as the complement of their cosine similarity:
\begin{equation}
    \mathcal{S}_{\text{LIM}}^{(l)} = 1 - \frac{X_{\text{in}}^{(l)} \cdot X_{\text{out}}^{(l)}}{\|X_{\text{in}}^{(l)}\| \|X_{\text{out}}^{(l)}\|}
\end{equation}
A higher cosine similarity implies that the layer's operations are highly redundant and induce minimal transformation. Therefore, a larger $\mathcal{S}_{\text{LIM}}^{(l)}$ indicates a more substantial feature transformation, which corresponds to higher layer sensitivity.

\paragraph{LSAQ.} 
Layer-Specific Adaptive Quantization (LSAQ) evaluates layer sensitivity by measuring the semantic transformation between the layer's input and output. Similar to LIM, it utilizes calibration data to obtain the hidden states $X_{\text{in}}^{(l)}$ and $X_{\text{out}}^{(l)}$. LSAQ projects these states onto the vocabulary space using the unembedding matrix $W_U$. It then constructs two sets, $C_{\text{in}}^{(l)}$ and $C_{\text{out}}^{(l)}$, comprising the top-$k$ most probable tokens decoded from the input and output:
\begin{equation}
\resizebox{\linewidth}{!}{%
$\displaystyle
C_{\text{in}}^{(l)} = f_{\text{top-}k}\!\left(X_{\text{in}}^{(l)} W_U\right), 
\quad 
C_{\text{out}}^{(l)} = f_{\text{top-}k}\!\left(X_{\text{out}}^{(l)} W_U\right)
$}
\end{equation}
The layer sensitivity score is computed as the complement of the Jaccard similarity between these two token sets:
\begin{equation}
    \mathcal{S}_{\text{LSAQ}}^{(l)} = 1 - \frac{|C_{\text{in}}^{(l)} \cap C_{\text{out}}^{(l)}|}{|C_{\text{in}}^{(l)} \cup C_{\text{out}}^{(l)}|}
\end{equation}
A smaller Jaccard similarity (resulting in a higher $\mathcal{S}_{\text{LSAQ}}^{(l)}$) signifies that the layer fundamentally alters the semantic representation of the input tokens. Consequently, layers with higher $\mathcal{S}_{\text{LSAQ}}^{(l)}$ scores are considered more sensitive and are prioritized for higher quantization precision.

\paragraph{LLM-MQ.}
LLM-MQ assesses layer sensitivity by approximating the perturbation in the model's loss function caused by quantization, utilizing a first-order Taylor expansion guided by a small calibration dataset. 
For a full-precision weight matrix $W \in \mathcal{W}_l$ and its corresponding gradient $G$ derived from the calibration data, the sensitivity metric with respect to a candidate bit-width $b$ is calculated as the absolute sum of the element-wise product between the gradient and the quantization error:
\begin{equation}
s_b(W) = \left| \sum G \odot (W - Q_b(W)) \right|
\end{equation}
where $Q_b(W)$ represents the $b$-bit quantized weight matrix, and $\odot$ denotes the element-wise multiplication. 

To obtain the layer-level sensitivity score, LLM-MQ computes the average sensitivity across all weight matrices within the layer. 
Thus, the sensitivity score for layer $l$ at bit-width $b$ is defined as:
\begin{equation}
\mathcal{S}_{\text{LLM-MQ}}^{(l)}(b) = \frac{1}{|\mathcal{W}_l|} \sum_{W \in \mathcal{W}_l} s_b(W)
\end{equation}
A larger $\mathcal{S}_{\text{LLM-MQ}}^{(l)}(b)$ score implies that quantizing this layer induces a more significant deviation in the loss function, indicating that the layer is highly sensitive and should be preserved in higher precision. 

\paragraph{LieQ.}
LieQ (Layer-wise information effectiveness Quantization) estimates layer sensitivity by analyzing the representational compactness, which measures the concentration of information in the layer's output representations. 
For a trained weight matrix $W \in \mathcal{W}_l$ and input hidden states $h$, it computes the projected activation $Z = Wh$, alongside a baseline $\tilde{Z} = \tilde{W}h$ generated by an untrained weight matrix $\tilde{W}$. 

LieQ performs Singular Value Decomposition (SVD) on the representations to obtain the normalized singular values $p_k = \frac{\sigma_k}{\sum_j \sigma_j}$. The representational compactness is then defined as the exponential of the Shannon entropy of this energy distribution:
\begin{equation}
Compact(Z) = \exp\left(-\sum_k p_k \log p_k\right)
\end{equation}

To quantify the structural shift induced by training, LieQ calculates the relative reduction in compactness for each weight matrix. 
The layer-level sensitivity score $\mathcal{S}_{\text{LieQ}}^{(l)}$ is formulated by averaging these relative changes across all weight matrices (e.g., query, key, and value projections) within the layer:
\begin{equation}
\mathcal{S}_{\text{LieQ}}^{(l)} = \frac{1}{|\mathcal{W}_l|} \sum_{W \in \mathcal{W}_l} \frac{Compact(\tilde{Z}) - Compact(Z)}{Compact(\tilde{Z})}
\end{equation}

A higher $\mathcal{S}_{\text{LieQ}}^{(l)}$ indicates that the layer has significantly concentrated its information into lower-rank manifolds during training, making its function highly irreplaceable and sensitive to quantization noise.

\subsection{Slim-LLM}
SliM-LLM (Salience-Driven Mixed-Precision Quantization) achieves low-bit compression by dynamically allocating group-wise bit-widths and calibrating quantizer parameters, utilizing GPTQ as its foundational backend. 
For a weight matrix $W \in \mathcal{W}_l$, it evaluates the salience of each weight element based on its magnitude and the input activation. The element-wise salience is mathematically approximated as $\delta_{i,j} \approx (w_{i,j} \cdot \|x_j\|_2)^2$, where $x_j$ represents the corresponding activation channel.

To perform Salience-Determined Bit Allocation (SBA), SliM-LLM computes the average salience for each structured group within the weight matrix. It then assigns different bit-widths (e.g., 2-bit or 3-bit) to these groups by minimizing the Kullback-Leibler (KL) divergence between the outputs of the full-precision and quantized layers, while strictly maintaining the target average bit-width budget.

Furthermore, to protect highly critical weights within each group, SliM-LLM introduces Salience-Weighted Quantizer Calibration (SQC). It isolates locally salient weights—typically outliers beyond a $3\sigma$ threshold and calibrates the quantizer's scale and zero-point by minimizing a weighted reconstruction loss. 
Finally, the optimized group quantizers operate within the GPTQ framework, which sequentially quantizes the weights and compensates for the residual quantization errors using the inverse Hessian matrix.

%% file: appendix/appendx_PTQ_backends.tex
\paragraph{HQQ.} 
Half-Quadratic Quantization (HQQ) is a highly efficient, calibration-free quantization method that optimizes the quantization parameters by minimizing a sparsity-promoting $\ell_{p<1}$-norm of the quantization error. For a full-precision weight matrix $W \in \mathcal{W}_l$, HQQ aims to find the optimal scaling factor $s$ and zero-point $z$ by formulating the objective as:
\begin{equation}
\arg\min_{z,s} \phi\left(W - Q_{z,s}^{-1}(Q_{z,s}(W))\right)
\end{equation}
where $\phi(\cdot)$ denotes the $\ell_{p<1}$-norm loss. Unlike the conventional squared error, this non-convex loss function effectively models and captures the heavy-tailed distribution of weight outliers through a hyper-Laplacian distribution. The quantization and de-quantization operators are defined as $Q_{z,s}(W) = \text{round}(W/s + z)$ and $Q_{z,s}^{-1}(W_q) = s(W_q - z)$, respectively.

To tackle this non-convex optimization problem, HQQ employs a Half-Quadratic solver by introducing an auxiliary continuous variable $W_e$. By fixing the scaling parameter $s$, the optimization of $z$ is reformulated as:
\begin{equation}
\resizebox{\linewidth}{!}{%
$\displaystyle
\begin{aligned}
\arg\min_{z, W_e} \phi(W_e) + \frac{\beta}{2} \left\|W_e - \left(W - Q_{z}^{-1}(Q_{z}(W))\right)\right\|_2^2
\end{aligned}$%
}
\end{equation}
where $\beta$ is a penalty parameter. This formulation allows the main problem to be decomposed into two tractable sub-problems that are efficiently solved via alternate optimization. Because HQQ relies exclusively on the weights without requiring any calibration data, it successfully avoids potential over-fitting and serves as an architecture-agnostic PTQ backend.

\paragraph{GPTQ.} 
GPTQ is a highly efficient calibration-based quantization method that builds upon the Optimal Brain Quantizer (OBQ) framework. For a full-precision weight matrix $W \in \mathcal{W}_l$ and calibration input activations $X$, GPTQ formulates the quantization process as a layer-wise reconstruction problem. It aims to minimize the output squared error between the full-precision and quantized layer:
\begin{equation}
\arg\min_{\hat{W}} \| WX - \hat{W}X \|_2^2
\end{equation}
where $\hat{W}$ represents the quantized weight matrix. 

To mitigate the performance degradation caused by precision reduction, GPTQ actively compensates for the quantization error of each weight. It does so by updating the remaining unquantized weights based on the inverse Hessian matrix, $H^{-1}$, of the input activations. While the original OBQ method quantizes weights individually and strictly requires recomputing $H^{-1}$ at each step, GPTQ innovates by quantizing the weights column-wise. 

This column-wise quantization strategy successfully eliminates the expensive repeated calculations of the inverse Hessian matrix, making the optimization highly tractable. Consequently, GPTQ can smoothly scale to massive language models with hundreds of billions of parameters, while concurrently accelerating the generation speed through its heavily optimized mixed-precision matrix multiplication kernels.

%% file: appendix/appendix_further_analysis.tex
\input{tabs/larger_scale_table_detailed}

\begin{figure}[!th]
    \centering \centerline{\includegraphics[width=\columnwidth]{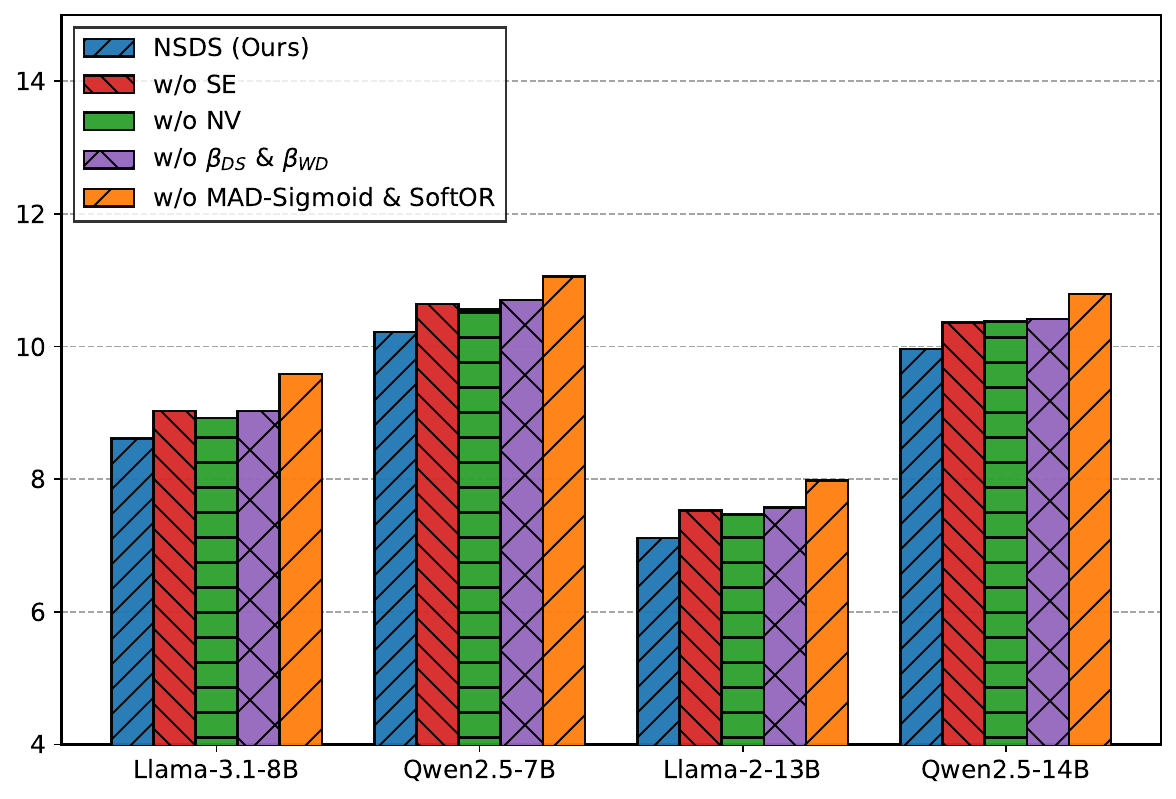}}
    \vspace{-0.2cm}
    \caption{Average perplexity results of the ablation analysis on \nsds across all language modeling benchmarks. ``w/o'' denotes the exclusion of a specific component from \nsds during layer sensitivity estimation.}
    
\vspace{-0.2cm}
    \label{fig:ablation_ppl}
\end{figure}

\paragraph{Language Modeling Results of Ablation Analysis.}
To comprehensively understand the impact of each component within \nsds, we further evaluate the ablation variants on language modeling benchmarks (WikiText-2 and C4). 
As illustrated in Fig.~\ref{fig:ablation_ppl}, the average perplexity results exhibit a degradation trend that aligns with the language reasoning accuracy shown in Fig.~\ref{fig:ablation_accuracy}. 
Specifically, the ``w/o \madsigmoid \& \softor'' variant consistently yields the most severe PPL increase, reaffirming that proper normalization and aggregation are crucial for obtaining the final layer sensitivity scores. 
Similarly, omitting \seshort or \nvshort causes a noticeable PPL surge. The consistent deterioration across all ablation settings on both reasoning and language modeling tasks validates the necessity of our joint numerical-structural design and the refined aggregation strategy.

\begin{figure}[!th]
    \centering \centerline{\includegraphics[width=\columnwidth]{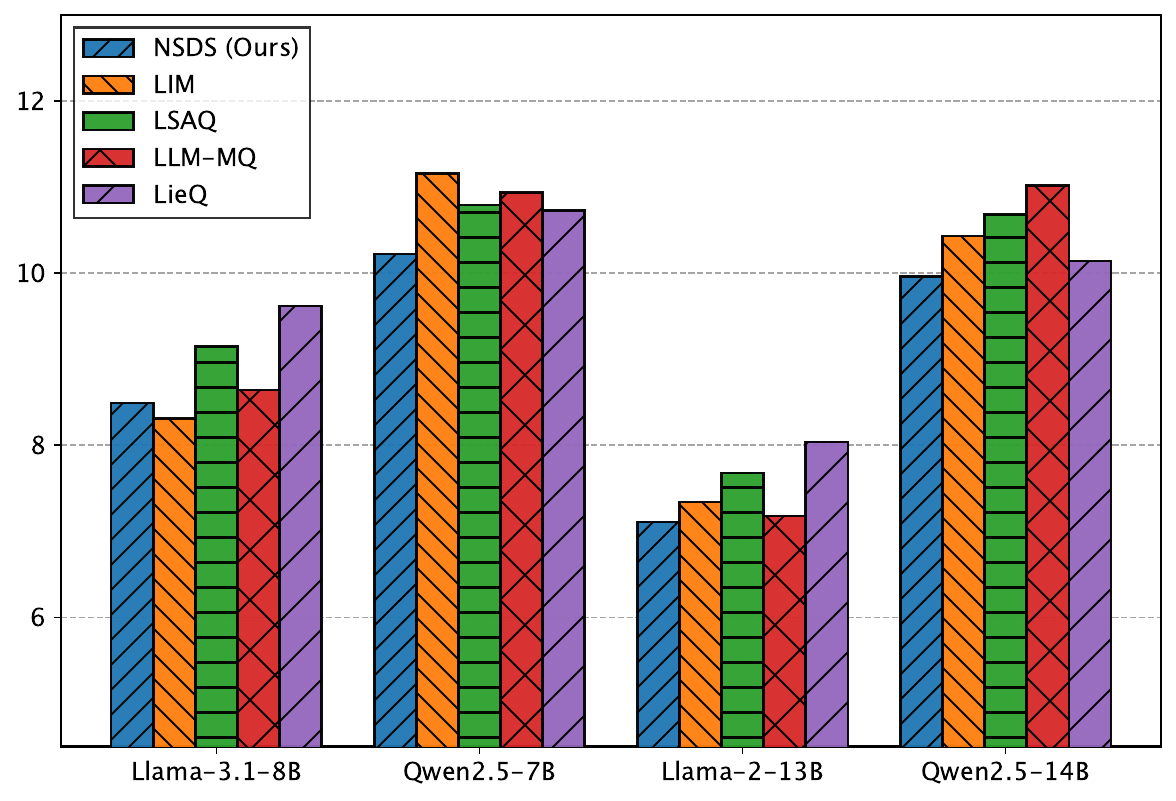}}
    \vspace{-0.2cm}
    \caption{Average perplexity comparison between \nsds framework and calibration-based baselines across generic reasoning benchmarks.}
    \vspace{-0.4cm}
\label{fig:calib_comparison_ppl}
\end{figure}

\paragraph{Language Modeling Results of Calibration-based Methods.}
To further validate the robustness of our data-free \nsds framework, we compare its average PPL on WikiText-2 and C4 with the calibration-based methods in Fig.~\ref{fig:calib_comparison_ppl}.
The results align with the accuracy results in Fig.~\ref{fig:calib_comparison_acc}.
Specifically, \nsds consistently maintains a top-two ranking with low PPL across all four LLMs.
Meanwhile, calibration-based baselines such as LIM and LLM-MQ show fluctuating PPL scores that depend on the specific target model.

\begin{figure}[!th]
    \centering \centerline{\includegraphics[width=\columnwidth]{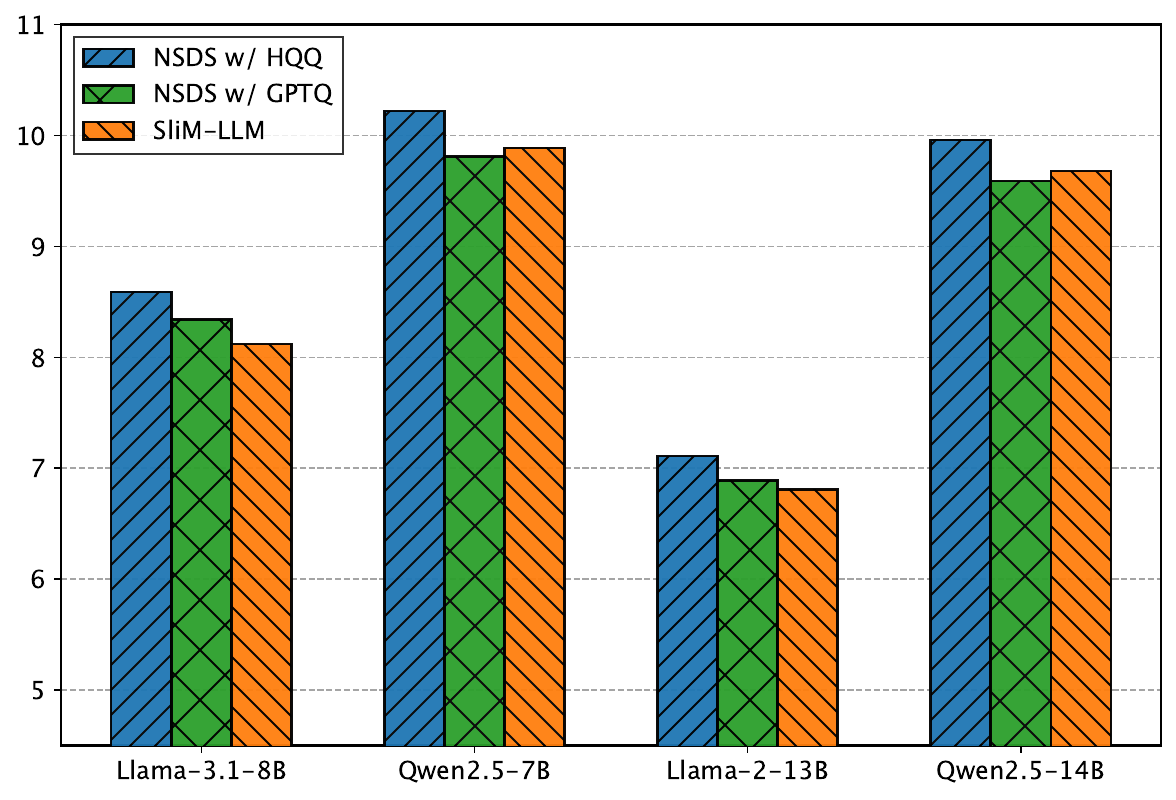}}
    \vspace{-0.2cm}
    \caption{Average perplexity on language modeling benchmarks when integrating \nsds with different PTQ backends, compared against the SliM-LLM.}
    \vspace{-0.2cm}
\label{fig:integration_ppl}
\end{figure}

\paragraph{Language Modeling Results of Integration with Other PTQ Backends.}
We report the average PPL results of \nsds integrated with different PTQ backends, along with the SliM-LLM baseline in Fig.~\ref{fig:integration_ppl}.
The results exhibit a trend similar to the accuracy in Fig.~\ref{fig:integration_acc}.
Upgrading the backend to GPTQ yields lower PPL across all evaluated models.
This indicates that the orthogonality of \nsds holds true for language modeling tasks as well, where upgrading to a stronger PTQ backend yields better performance.

%% file: tabs/larger_scale_table_detailed.tex
\begin{table*}[!th]
    \centering

    \renewcommand{\arraystretch}{1.2}  
    \fontsize{10}{8}\selectfont      
    \setlength{\tabcolsep}{5pt}      

    \begin{tabular}{l cccccc cc}
        \toprule[1.5pt]
        & \multicolumn{6}{c}{Language Reasoning ($\uparrow$)} & \multicolumn{2}{c}{Language Modeling ($\downarrow$)} \\
        \cmidrule(lr){2-7} \cmidrule(lr){8-9}
         & ARC-C & Hellaswag & PIQA & BoolQ & Winogrande & TruthfulQA & Wikitext-2 & C4 \\
        \midrule

        \rowcolor{gray!20} & \multicolumn{8}{c}{\textbf{Llama-2-13B}} \\ \addlinespace[1pt]
        FP16 & 59.47 & 82.22 & 79.05 & 80.55 & 76.16 & 25.95 & 4.88 & 6.47 \\ 
        \addlinespace[1pt] \hdashline[1pt/1pt] \addlinespace[2pt] 
        MSE & 46.71 & 73.27 & 73.35 & 72.58 & 71.63 & 21.96 & 8.55 & 10.21 \\
        EWQ & 48.14 & 75.11 & 75.38 & 75.22 & 73.43 & 22.85 & 6.77 & 8.43 \\
        ZD & 47.89 & 74.34 & 74.83 & 74.01 & 72.69 & 22.23 & 7.08 & 8.89 \\
        KurtBoost & 48.54 & 75.95 & 75.97 & 75.91 & 73.12 & 23.04 & 6.62 & 8.41 \\
        NSDS (Ours) & \textbf{49.74} & \textbf{76.86} & \textbf{76.78} & \textbf{76.75} & \textbf{73.96} & \textbf{23.86} & \textbf{6.23} & \textbf{7.99} \\
        \midrule
        
        \rowcolor{gray!20} & \multicolumn{8}{c}{\textbf{Qwen2.5-14B}} \\ \addlinespace[1pt]
        FP16 & 67.24 & 84.34 & 81.07 & 85.23 & 81.45 & 40.15 & 5.29 & 9.15 \\
        \addlinespace[1pt] \hdashline[1pt/1pt] \addlinespace[2pt] 
        MSE & 55.94 & 75.37 & 74.81 & 76.94 & 73.05 & 29.63 & 8.93 & 12.31 \\
        EWQ & 55.03 & 74.52 & 73.95 & 75.69 & 72.58 & 28.44 & 9.37 & 12.88 \\
        ZD & 57.38 & 76.49 & 76.32 & 77.97 & 73.91 & 30.45 & 8.54 & 11.97 \\
        Kurtosis & 56.76 & 76.58 & 75.49 & 78.13 & 73.06 & 30.77 & 8.46 & 12.36 \\
        NSDS (Ours) & \textbf{58.27} & \textbf{77.34} & \textbf{76.78} & \textbf{78.62} & \textbf{74.28} & \textbf{31.58} & \textbf{8.28} & \textbf{11.65} \\
        
        \bottomrule[1.5pt]
    \end{tabular}
    \vspace{-0.2cm}
    \caption{Results of baseline methods and our \nsds on Llama-2-13B and Qwen2.5-14B, evaluated on six language reasoning and two language modeling benchmarks.}
    \label{tab:larger_main_results}
    \vspace{-0.4cm}
\end{table*}